\pdfoutput=1

\documentclass[11pt]{article}

\usepackage[final]{acl}

\usepackage{times}
\usepackage{latexsym}

\usepackage[T1]{fontenc}

\usepackage[utf8]{inputenc}

\usepackage{microtype}

\usepackage{inconsolata}

\usepackage{graphicx}

\usepackage{kotex}
\usepackage{amsmath}
\usepackage{enumitem}
\usepackage{comment}
\usepackage{physics}
\usepackage{anyfontsize}
\usepackage{svg}
\usepackage{multirow}
\usepackage{multicol}
\usepackage{array}
\usepackage{booktabs}
\usepackage[normalem]{ulem}
\useunder{\uline}{\ul}{}
\usepackage{bbm}
\usepackage{algorithm}
\usepackage{algpseudocode}
\usepackage{rotating}
\usepackage{multirow}
\usepackage{xcolor, colortbl}
\usepackage{pifont}
\usepackage{amssymb}
\usepackage{makecell}

\usepackage{amsthm}
\newtheorem*{remark}{Theorem}

\usepackage[capitalize]{cleveref}
\crefname{section}{Sec.}{Secs.}
\Crefname{section}{Section}{Sections}
\Crefname{table}{Table}{Tables}
\crefname{table}{Tab.}{Tabs.}

\title{AMQ: Enabling AutoML for Mixed-precision Weight-Only Quantization of Large Language Models}

\author{
 \textbf{Sangjun Lee}\textsuperscript{1}$^*$ \quad
 \textbf{Seung-taek Woo}\textsuperscript{1}$^*$ \quad
 \textbf{Jungyu Jin}\textsuperscript{1} \quad
 \textbf{Changhun Lee}\textsuperscript{2} \quad
 \textbf{Eunhyeok Park}\textsuperscript{1} \quad \\
 \textsuperscript{1} Graduate School of Artificial Intelligence \\
 \textsuperscript{2} Department of Convergence IT Engineering \\
 Pohang University of Science and Technology (POSTECH) \\
 \texttt{\{leesangjun, wst9909, jgjin0317, changhun.lee, eh.park\}@postech.ac.kr}
}

\begin{document}
\maketitle

\begin{abstract}
To enable broader deployment of Large Language Models (LLMs), it is essential to identify the best-performing model under strict memory constraints. We present AMQ, Automated Mixed-Precision Weight-Only Quantization, a framework that assigns layer-wise quantization bit-widths to optimally balance model quality and memory usage. However, the combinatorial search space, with over $10^{100}$ possible configurations, makes conventional black-box optimization infeasible. AMQ overcomes this challenge through four key innovations: (1) \textbf{search space pruning} using prior knowledge to exclude unpromising configurations, (2) \textbf{quantization proxy} to bypass costly format conversions during search, (3) \textbf{quality predictor} to minimize evaluation overhead, and (4) \textbf{iterative search-and-update} strategy for fast and stable convergence. By integrating these components, AMQ efficiently explores the quality–efficiency landscape, reaching the Pareto frontier and yielding LLMs that are both compact and high-performing. Our code is available at \url{https://github.com/dlwns147/amq}.
\end{abstract}

\begingroup
\renewcommand\thefootnote{*}
\footnotetext{These authors contributed equally.}
\endgroup

\section{Introduction}
Weight-only quantization is a powerful optimization technique for large language models (LLMs), significantly reducing memory usage and alleviating performance bottlenecks by lowering the bit-width of model weights. Recent advances~\cite{gptq,awq,QUIP} have demonstrated that even 4-bit quantization can preserve model accuracy. However, pushing below 4 bits often leads to substantial accuracy degradation due to increased quantization error. This raises a critical question: \textbf{“Given a fixed memory budget, how can we compress an LLM to achieve the best possible performance?”}

\begin{figure}[btp]
    \centering
    \includegraphics[width=0.95\linewidth]{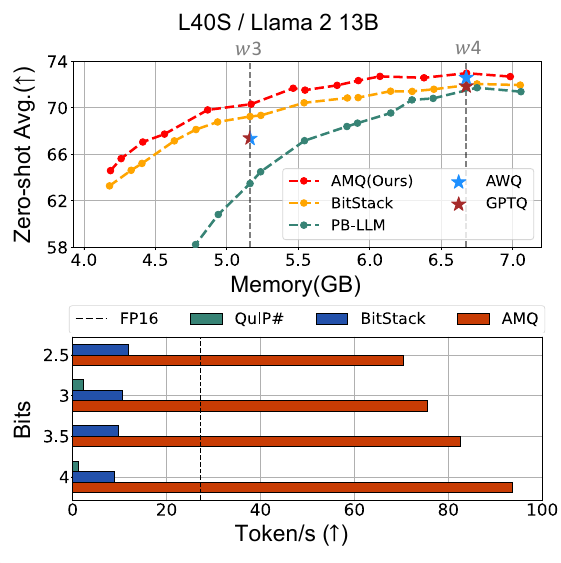}
    \vspace{-0.5cm}
    \caption{(Top): Trade-off between memory usage and average zero-shot accuracy on ARC-Easy, ARC-Challenge, PIQA, HellaSwag, WinoGrande, and BoolQ. (Bottom): Inference speed. 
    See \Cref{sec:expr} and Appendix~\ref{detailed_expr_setting} for more details. 
    }
    \vspace{-0.6cm}
    \label{fig:figure_1}
\end{figure}

Building on this motivation, our study investigates the Pareto frontier of weight-only quantization to maximize model quality under a fixed memory budget. A key observation is that quantization sensitivity varies widely across models, modules, and layers, indicating that fine-grained bit-width allocation can open up a new possible solution, enhancing the quality-efficiency trade-off. However, this flexibility comes at a cost: it dramatically enlarges the configuration space, posing a key yet largely unexplored challenge, 
\textbf{how to efficiently identify the optimal configuration within such a vast search space}.

This problem can be formulated as a discrete combinatorial optimization task, which is known to be NP-complete. A common strategy is to apply black-box optimization methods such as simulated annealing or genetic algorithms~\cite{simulated_annealing, genetic_algorithm}. However, these methods are impractical in our context. Due to the massive scale of LLMs, even converting between different bit-width formats can take several hours. Moreover, evaluating model quality on a full validation set is computationally expensive, often requiring thousands to millions of iterations per configuration. Consequently, despite aggressive evaluation pruning, black-box approaches remain prohibitively costly and unsuitable for real-world deployment.

To address these limitations, we propose \textbf{AMQ}, Automated Mixed-Precision Weight-Only Quantization, a novel framework that automatically identifies the optimal quantization configuration for any given model, maximizing accuracy and execution efficiency within a fixed memory budget, without requiring expert intervention.
AMQ is built upon four key innovations: (1) \textbf{search space pruning} based on prior knowledge to eliminate unpromising configurations, (2) \textbf{quantization proxy} to avoid costly bit-width format conversions during the search,
(3) \textbf{quality predictor} to reduce evaluation overhead, and
(4) \textbf{iterative search-and-update} strategy that enables fast and stable convergence. Together, these techniques make black-box optimization tractable, allowing AMQ to discover near-optimal configurations for fine-grained LLMs quantization. Extensive experiments show that AMQ consistently outperforms existing methods in accuracy under the same memory constraints, while also delivering the highest throughput, as illustrated in Figure~\ref{fig:figure_1}.

\section{Related Works}

\subsection{Weight-only Quantization}
Weight-only quantization for LLMs reduces model memory usage, addressing hardware constraints and improving inference speed. GPTQ \cite{gptq} and AWQ \cite{awq} successfully minimized performance degradation while reducing model weights to 3-bit or 4-bit precision. However, these approaches rely on fixed-precision quantization, making them less adaptable to diverse memory environments.
The advanced mixed-precision quantization \cite{billm, PB-LLM, owq, slim_llm, SqueezeLLM, qeft}, which assigns different bit precisions to specific groups or channels within linear layers based on weight distribution, has further improved performance. However, these methods require irregular data access due to mixed precision, making it challenging to achieve actual speedup in real-world deployment.

\subsection{Model Compression with Varying Targets}
Recent research has focused on model compression techniques that allow LLMs to be flexibly adjusted to accommodate various memory restrictions. LLM-MQ \cite{llm_mq} and CMPQ \cite{cmpq} perform mixed-precision quantization at the linear or channel level to match variable target bits, while sharing the common problem of irregular pattern from mixed precision. BitStack \cite{bitstack} stores weights using low-rank decomposition and reconstructs them by loading residual blocks within memory constraints. It is an alternate approach supporting various compression targets, but the weight reconstruction process slows down inference notably. 

\subsection{AutoML for Neural Architecture Search}
Neural architecture search (NAS) is a prominent branch of research within the field of AutoML, offering various solutions to combinatorial optimization problems, which align closely with our own research. To identify neural network architectures within a discrete space that optimize accuracy and performance, numerous studies have explored black-box optimization methods and differentiable approaches \cite{rlnas, paramefficientnas, regularized_nas, efficientnet}. Among these, we drew significant inspiration from research based on neural predictors \cite{neural_predictor, fbnetv2, powerful_predictor}. In particular, the quality predictor serves as a key common element that substantially reduces search costs. However, our unique contribution lies in additional ideas that consider the characteristics of quantization, which is also highly important to make AMQ feasible.

\section{Method}

\begin{figure}[tb!]
    \centering
    \includegraphics[width=\linewidth]{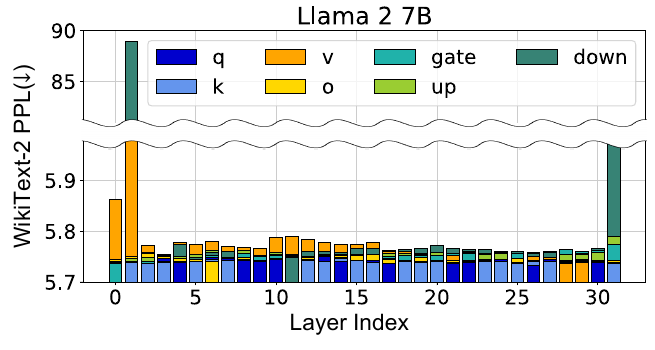}
    \vspace{-0.9cm}
    \caption{Quantization sensitivity of Llama 2 7B per linear layer with HQQ. Result is the perplexity on the WikiText-2 test set.}
    \vspace{-0.5cm}
    \label{fig:sensitivity}
\end{figure}


The primary objective of this work is to make the search for optimal quantization configurations computationally feasible. To effectively navigate the trade-off between memory and accuracy, we aim to construct a Pareto frontier over various bit-width combinations. NSGA-II~\cite{nsga2}, a genetic algorithm-based method, is well-suited for this task, as it efficiently explores Pareto-optimal solutions across multiple objectives. If search over the configuration space is tractable, one can simply select the most accurate model within the given memory budget from the resulting Pareto frontier. However, in practice, the search process requires hundreds of thousands of quantization/evaluation runs, resulting in prohibitive costs. This section identifies the key sources of this overhead and introduces four core ideas designed to mitigate it.



\subsection{Search Space Design}

A key step in making the search efficient is to carefully define the configuration space. While finer-grained quantization (e.g., per-weight or per-channel) may offer more optimal configurations, it dramatically increases the number of meaningless candidates and raises the risk of suboptimal convergence. In addition, such granularity often breaks hardware-friendly memory access patterns, leading to significant slowdowns during inference.

\begin{figure}[]
    \centering    
    \includegraphics[width=\linewidth]{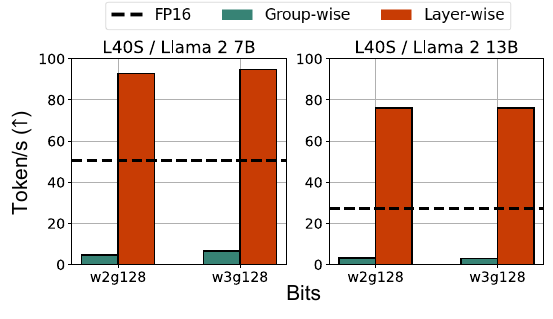}
    \vspace{-1cm}
    \caption{Llama 2 7/13B inference speed of FP16, group-wise mixed-precision quantization \cite{slim_llm} and layer-wise quantization on single NVIDIA L40S GPU.}
    \vspace{-0.4cm}
    \label{fig:group_linear_speed}
\end{figure}

To strike a balance between flexibility and efficiency, we adopt \textbf{layer-wise bit-width} as the core unit of our search space. This design aligns well with hardware execution and offers sufficient expressiveness for mixed-precision strategies. As shown in \Cref{fig:group_linear_speed}, Our empirical analysis shows that applying different bit-widths within a single linear layer ~\cite{slim_llm} introduces irregular memory access, causing considerable inference latency with only marginal gains in quality. 
Instead, assigning a single bit-width, chosen from 2, 3, or 4 bits, to each linear layer, using grouped quantization~\cite{awq}, provides a good trade-off: it preserves high-quality representations while maintaining a rich search space. For example, Llama 2 7B~\cite{llama2} has 224 linear layers, yielding a search space size of $3^{224} \approx 10^{106}$ possibilities. For the group size of 128, the range of our search space is $[2.25, 4.25]$.

\subsection{Space Pruning via Prior Knowledge}
However, the vast configuration space contains many low-quality candidates that are unlikely to contribute to an optimal solution. Our analysis reveals substantial variation in quantization sensitivity across linear layers. As shown in Figure~\ref{fig:sensitivity}, individually quantizing a single layer to 2-bit, while keeping all others at 4-bit, results in widely varying degrees of perplexity degradation. This suggests that some layers are highly sensitive and should remain at higher precision (e.g., 4-bit), while others are robust enough to tolerate lower bit-widths.

Based on this observation, we propose a simple yet effective strategy to refine the search space by pruning configurations unlikely to yield high-quality results. However, overly aggressive pruning risks eliminating promising candidates. To mitigate this, we adopt a conservative filtering criterion. We first measure per-layer sensitivity using a small calibration set, following the aforementioned procedure. Layers with sensitivity exceeding twice the median are treated as outliers and fixed to 4-bit precision; the remaining layers define the active search space. This model-aware pruning approach substantially reduces search complexity while preserving flexibility. For Llama 2 7B, 13B, and 70B, only 4, 3, and 5 layers are excluded, corresponding to just 1.8\%, 1.1\%, and 0.9\% of all linear layers, respectively. Despite its simplicity, this modification leads to a notable gain, as demonstrated in Section~\ref{sec:prune}.

\begin{figure}[t!]
    \centering    
    \includegraphics[width=\linewidth]{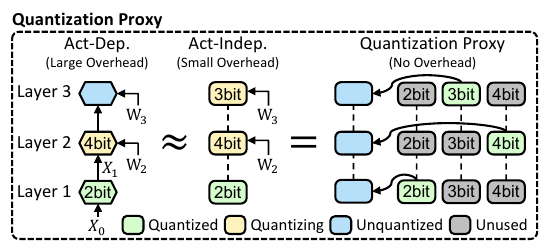}
    \vspace{-0.8cm}
    \caption{Illustration of activation-dependent vs. activation-independent quantization and Quantization Proxy.}
    \vspace{-0.6cm}
    \label{fig:quantization_proxy}
\end{figure}

\begin{figure*}[tb!]
    \centering
    \includegraphics[width=0.9\linewidth]{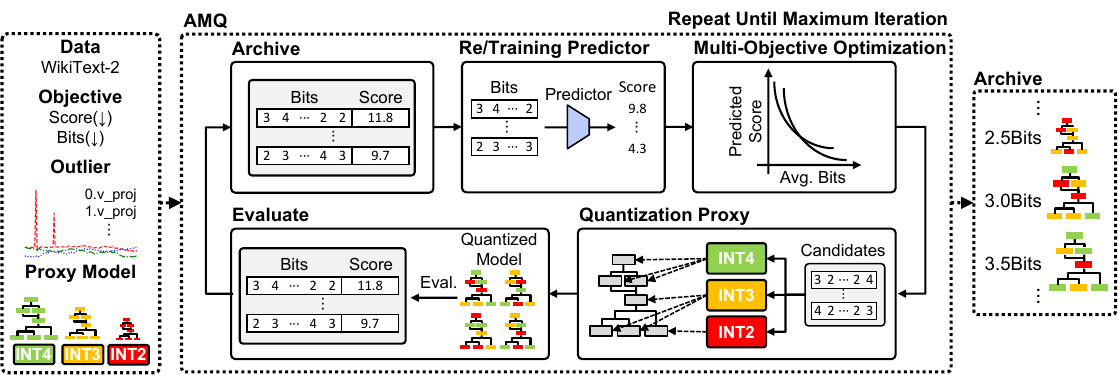} 
    \vspace{-0.2cm}
    \caption{The overview of our AMQ pipeline.}
    \label{fig:amq_overview}
    \vspace{-0.7cm}
\end{figure*}

\subsection{Quantization Proxy}


One of the key challenges in layer-wise mixed-precision search is the high computational cost of generating quantized model representations. For example, the widely used AWQ~\cite{awq} takes approximately 1.5 hours on a single A100 GPU to convert the FP16 weights of a 70B model into 4-bit precision. Additionally, AWQ employs an activation-dependent quantization scheme in which each layer’s quantization depends on the activation results of preceding layers. Consequently, every distinct bit-width configuration requires a separate, end-to-end conversion process.




While activation-dependent quantization techniques achieve state-of-the-art model quality, their computational cost makes them impractical for direct use within a search process. To address this limitation, we propose a \textbf{quantization proxy}, a lightweight approximation that assembles a quantized model by selecting precomputed versions of each linear layer based on the desired bit-width configuration, as illustrated in \Cref{fig:quantization_proxy}. The motivation behind this approach is formalized in the following theorem:

\begin{remark} \label{thm:pareto_frontier_preservation}

    Let $X$ be the bit-width search space with size function $S:X\to\mathbb{R}_{>0}$ and injective model-quality scores $Q_1,Q_2:X\to\mathbb{R}$, where $Q_1$ corresponds to proxy quantization and $Q_2$ to activation-dependent quantization. If
    $$Q_1(x)<Q_1(y)\;\Rightarrow\; Q_2(x)<Q_2(y)$$    
    for all $x,y\in X$, then the Pareto frontier for $(Q_1,S)$ coincides with that for $(Q_2,S)$.
\end{remark}

\begin{figure}[t]
    \centering    
    \includegraphics[width=\linewidth]{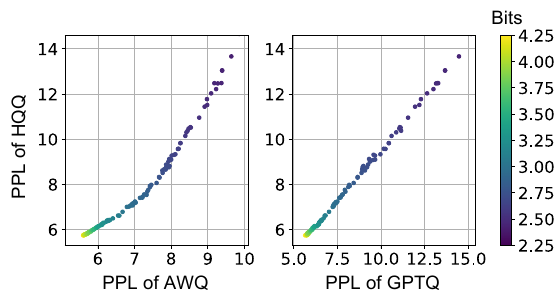}
    \vspace{-0.9cm}
    \caption{WikiText-2 perplexity (PPL) of HQQ, asymmetric clipping AWQ, and GPTQ on Llama 2 7B, evaluated over a randomly sampled 20\% of Pareto frontiers identified by AMQ.
    }
    \vspace{-0.4cm}
    \label{fig:hqq_awq_gptq_pareto_front}
\end{figure}

The proof of the theorem is provided in Appendix~\ref{proof_pareto_frontier}. To apply this proxy-based approach, we evaluate several existing quantization methods and identify HQQ~\cite{hqq}, an activation-independent technique that quantizes weights without requiring activation data, as a strong candidate. As shown in Figure~\ref{fig:hqq_awq_gptq_pareto_front}, the Pareto frontier obtained using HQQ closely aligns with those derived from activation-dependent methods such as GPTQ and asymmetric clipping AWQ~\cite{LLMC}. This implies that we can search for optimal configurations using HQQ, and then transfer the resulting bit-width assignments to GPTQ or AWQ for final deployment. To leverage the activation-independent nature of HQQ, we precompute each linear layer at 2-, 3-, and 4-bit precision. Given any target configuration, a quantized model can be efficiently assembled by selecting the corresponding precomputed layers. This approach significantly reduces computational overhead while preserving the quality of solutions discovered during the search.

\subsection{Quality Predictor}
\label{quality_score_predictor}

To efficiently guide the search, we assess model quality using logit similarity, based on the intuition that a quantized model retains quality if its output distribution remains close to that of the original. 
Specifically, we measure the Jensen–Shannon divergence (JSD) between the logits of the quantized model and that of the original model.

Although computing JSD requires only a forward pass, evaluating each quantized model over the full dataset remains expensive, especially when scaling to thousands of configurations. To mitigate this overhead, we introduce a surrogate quality predictor that estimates model performance using a small set of JSD-labeled samples. We adopt a Radial Basis Function (RBF) model~\cite{rbf}, which predicts the expected quality of unseen configurations during the search. Leveraging this model, we can rapidly evaluate hundreds of thousands of configurations. However, since the predictions are approximate, the resulting Pareto frontier may contain errors. These inaccuracies are mitigated through the iterative search-and-update process described below, ensuring that model quality is preserved despite the use of the predictor.

\subsection{Iterative Search-and-update}


Building on the proposed quantization proxy and quality predictor, we apply NSGA-II to search for optimal bit-width configurations. Figure~\ref{fig:amq_overview} illustrates the complete AMQ pipeline, and the corresponding pseudo-code is provided in Algorithm~\ref{alg:search}.

The process begins with search space pruning to eliminate unpromising candidates. We then perform random sampling, evaluating hundreds of configurations to initialize the archive, a set of observed samples, which is used to train the initial quality predictor. Next, NSGA-II explores the Pareto front, optimizing two objectives: the predicted model quality (using the predictor) and the average bit-width (as a proxy for memory usage). 

The candidate solutions found by NSGA-II are then evaluated to obtain their true quality scores, which are used to update the archive. The quality predictor is retrained on this updated archive, and the search process is repeated for a fixed number of iterations. Finally, the best configuration that satisfies the memory constraint is selected.

This iterative pipeline ensures that the predictor is continuously refined using high-quality samples discovered during the search. As a result, both performance estimation and solution quality improve over time, enabling efficient exploration of the quantization search space.

\section{Experiments}
\label{sec:expr}

\renewcommand{\arraystretch}{0.8}
\begin{table*}[ht]
\setlength{\tabcolsep}{1mm}
\resizebox{\textwidth}{!}{%
\begin{tabular}{c|c|c|c||cc||ccccccc}
\midrule
\textbf{\large Model} & \textbf{\begin{tabular}[c]{@{}c@{}}\textbf{\large{Mem.}}\\ \textbf{\large{(MB)}}\end{tabular}} & \textbf{\begin{tabular}[c]{@{}c@{}}\textbf{\large{Avg.}}\\ \textbf{\large{Bits}}\end{tabular}} &  \textbf{\large Method} &  \textbf{\large Wiki2($\downarrow$)} &  \textbf{\large C4($\downarrow$)} &  \textbf{\large ARC-e($\uparrow$)} &  \textbf{\large ARC-c($\uparrow$)} &  \textbf{\large PIQA($\uparrow$)} &  \textbf{\large HellaS.($\uparrow$)} &  \textbf{\large WinoG.($\uparrow$)} & \textbf{\large BoolQ($\uparrow$)} & \textbf{\large Avg.($\uparrow$)} \\ \midrule
\multirow{12}{*}{\large \textbf{7B}} & 12,853 & 16 & FP16 & 5.47 & 7.26 & 74.58 & 46.25 & 79.11 & 76.00 & 69.22 & 77.77 & 70.49 \\ \cmidrule{2-13}
 & \multirow{3}{*}{2,431} & \multirow{3}{*}{2.5} & PB-LLM & 24.53 & 32.05 & 37.50 & 23.04 & 58.00 & 34.25 & 51.85 & 61.77 & 44.40 \\
 & &  & BitStack & \textbf{8.92} & \textbf{12.09} & 55.56 & 33.62 & 72.31 & 61.85 & \textbf{63.77} & \textbf{72.35} & 59.91 \\
 \rowcolor{gray!25} \cellcolor{white} & \cellcolor{white} & \cellcolor{white} & AMQ & 9.24 & 12.37 & \textbf{58.88} & \textbf{36.86} & \textbf{73.50} & \textbf{65.01} & 62.75 & 66.39 & \textbf{60.56} \\ \cmidrule{2-13}
 & \multirow{3}{*}{2,817} & \multirow{3}{*}{3.0} & PB-LLM & 11.60 & 14.81 & 53.20 & 29.10 & 70.02 & 53.82 & 61.72 & 71.31 & 56.53 \\
 & &  & BitStack & 7.46 & 10.13 & 62.16 & 37.63 & 74.81 & 66.96 & 66.38 & \textbf{73.82} & 63.63 \\
 \rowcolor{gray!25} \cellcolor{white} & \cellcolor{white} & \cellcolor{white} & AMQ & \textbf{6.83} & \textbf{9.03} & \textbf{68.22} & \textbf{41.72} & \textbf{76.55} & \textbf{71.27} & \textbf{67.32} & 68.44 & \textbf{65.59} \\ \cmidrule{2-13}
 & \multirow{3}{*}{3,203} & \multirow{3}{*}{3.5} & PB-LLM & 7.90 & 10.40 & 62.75 & 36.60 & 74.92 & 65.43 & 67.80 & \textbf{77.25} & 64.12 \\
 & &  & BitStack & 6.72 & 9.04 & 64.06 & 40.44 & 76.17 & 69.61 & 67.88 & 75.11 & 65.54 \\
 \rowcolor{gray!25} \cellcolor{white} & \cellcolor{white} & \cellcolor{white} & AMQ & \textbf{5.95} & \textbf{7.90} & \textbf{71.55} & \textbf{44.20} & \textbf{77.86} & \textbf{73.92} & \textbf{69.06} & 73.88 & \textbf{68.41} \\ \midrule
\multirow{12}{*}{\large \textbf{13B}} & 24,826 & 16 & FP16 & 4.88 & 6.73 & 77.53 & 49.15 & 80.52 & 79.37 & 72.30 & 80.55 & 73.23 \\ \cmidrule{2-13}
 & \multirow{3}{*}{4,408} & \multirow{3}{*}{2.5} & PB-LLM & 32.70 & 42.50 & 42.05 & 24.15 & 61.48 & 35.95 & 53.28 & 62.39 & 46.55 \\
 & &  & BitStack & 7.46 & 10.13 & \textbf{67.89} & 38.23 & 75.73 & 66.89 & 67.25 & 75.26 & 65.21 \\
 \rowcolor{gray!25} \cellcolor{white} & \cellcolor{white} & \cellcolor{white} & AMQ & \textbf{6.88} & \textbf{9.46} & 67.38 & \textbf{40.19} & \textbf{77.09} & \textbf{71.11} & \textbf{69.38} & \textbf{77.16} & \textbf{67.05} \\ \cmidrule{2-13}
 & \multirow{3}{*}{5,164} & \multirow{3}{*}{3.0} & PB-LLM & 9.57 & 13.30 & 63.55 & 37.71 & 75.46 & 62.77 & 68.27 & 73.09 & 63.48 \\
 & &  & BitStack & 6.33 & 8.73 & \textbf{74.37} & 44.37 & 77.26 & 71.93 & 69.46 & 78.10 & 69.25 \\
 \rowcolor{gray!25} \cellcolor{white} & \cellcolor{white} & \cellcolor{white} & AMQ & \textbf{5.68} & \textbf{7.80} & 72.18 & \textbf{45.39} & \textbf{78.35} & \textbf{76.02} & \textbf{70.32} & \textbf{79.57} & \textbf{70.31} \\ \cmidrule{2-13}
 & \multirow{3}{*}{5,920} & \multirow{3}{*}{3.5} & PB-LLM & 6.79 & 9.44 & 68.48 & 41.81 & 77.86 & 71.35 & 71.98 & \textbf{80.58} & 68.68 \\
 & &  & BitStack & 5.76 & 7.93 & 75.67 & 45.56 & 78.62 & 73.61 & 71.19 & \textbf{80.58} & 70.87 \\
 \rowcolor{gray!25} \cellcolor{white} & \cellcolor{white} & \cellcolor{white} & AMQ & \textbf{5.20} & \textbf{7.13} & \textbf{75.84} & \textbf{48.89} & \textbf{80.25} & \textbf{77.67} & \textbf{72.22} & 79.14 & \textbf{72.34} \\ \midrule
\multirow{9}{*}{\large \textbf{70B}} & 131,591 & 16 & FP16 & 3.32 & 5.71 & 81.02 & 57.25 & 82.70 & 83.78 & 77.98 & 83.79 & 77.75 \\ \cmidrule{2-13}
 && & BitStack & 4.91 & 7.43 & 76.73 & 51.11 & 79.82 & 77.25 & \textbf{75.93} & 76.70 & 72.92 \\
  \rowcolor{gray!25} \cellcolor{white} & \cellcolor{white}\multirow{-2}{*}{21,403} & \cellcolor{white}\multirow{-2}{*}{2.5} & AMQ & \textbf{4.90} & \textbf{7.25} & \textbf{77.02} & \textbf{51.28} & \textbf{80.03} & \textbf{77.84} & 75.61 & \textbf{80.73} & \textbf{73.75} \\ \cmidrule{2-13}
 & & & BitStack & 4.34 & 6.69 & \textbf{78.83} & 54.78 & 81.56 & 79.81 & \textbf{76.72} & 80.95 & 75.44   \\
 \rowcolor{gray!25} \cellcolor{white} & \cellcolor{white}\multirow{-2}{*}{25,483} & \cellcolor{white}\multirow{-2}{*}{3.0} & AMQ & \textbf{4.01} & \textbf{6.29} & 78.54 & \textbf{55.80} & \textbf{81.77} & \textbf{81.31} & 75.45 & \textbf{83.46} & \textbf{76.05} \\ \cmidrule{2-13}
 & & & BitStack & 3.95 & 6.30 & 79.71 & 56.23 & 81.99 & 81.06 & \textbf{77.51} & 82.48 & 76.50 \\
 \rowcolor{gray!25} \cellcolor{white} & \cellcolor{white}\multirow{-2}{*}{29,563} & \cellcolor{white}\multirow{-2}{*}{3.5} & AMQ & \textbf{3.62} & \textbf{5.93} & \textbf{79.80} & \textbf{57.68} & \textbf{82.21} & \textbf{82.80} & 77.19 & \textbf{82.97} & \textbf{77.11} \\ \bottomrule \bottomrule
\end{tabular}%
}
\vspace{-0.3cm}
\caption{Evaluation of Llama 2 7B/13B/70B models compressed by AMQ, BitStack and PB-LLM at average bit widths of 2.5, 3.0, and 3.5, showing WikiText-2 and C4 dataset perplexity (PPL) alongside zero-shot tasks accuracy. PB-LLM is excluded due to lack of 70B model support.}
\vspace{-0.4cm}
\label{tab:main_table}
\end{table*}


\renewcommand{\arraystretch}{1}

\begin{figure*}[bt!]
    \centering
    \includegraphics[width=\textwidth]{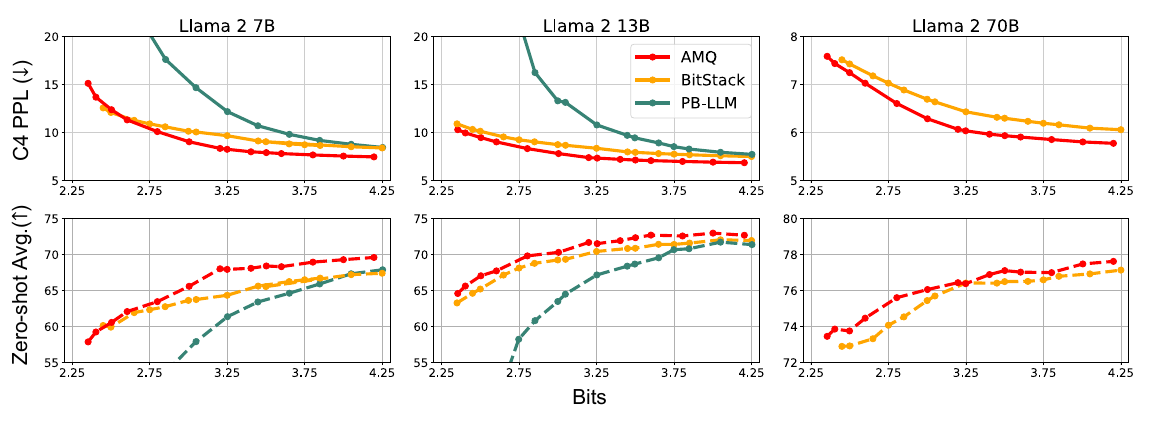}
    \vspace{-1cm}
    \caption{Trade-off between the average zero-shot accuracy and average bit-width over AMQ, BitStack and PB-LLM.}
    \vspace{-0.5cm}
    \label{fig:figure_5}
\end{figure*}

\begin{figure*}[bt]
    \centering
    \includegraphics[width=\textwidth]{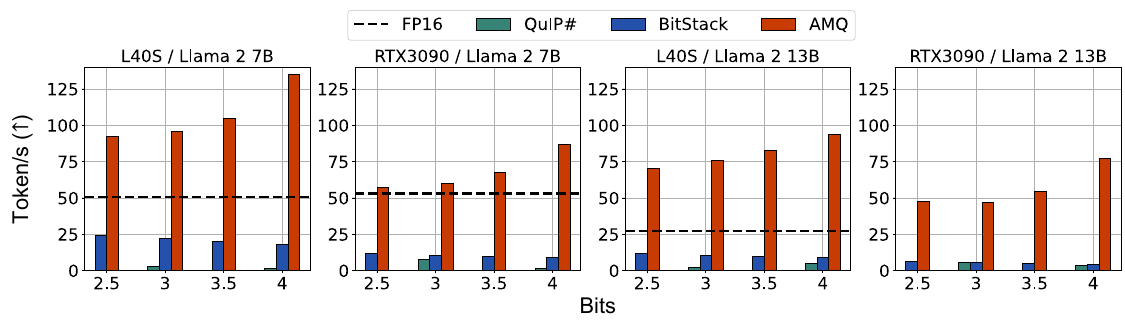}
    \vspace{-1cm}
    \caption{Inference speeds at various average bit-widths. Llama 2 13B (FP16) exceeds the VRAM capacity of a single RTX 3090.
    }
    \vspace{-0.5cm}
    \label{fig:Inference_speed}
\end{figure*}

\renewcommand{\arraystretch}{0.75}
\begin{table}[thb!]
\setlength{\tabcolsep}{1mm}
\resizebox{\columnwidth}{!}
{
\begin{tabular}{c|c|c|c||cc}
\midrule
\textbf{\large Model} & \textbf{\begin{tabular}[c]{@{}c@{}}\large Mem.\\ \large (MB)\end{tabular}} & \textbf{\begin{tabular}[c]{@{}c@{}}\large Avg.\\ \large Bits\end{tabular}} & \textbf{\large Method} & \textbf{\large MMLU} & \textbf{\large GSM8K} \\ \midrule
 & 13,825 & 16 & FP16 & 62.40 & 36.32 \\ \cmidrule{2-6}
 & & & BitStack & 34.60 & 3.18 \\
 \rowcolor{gray!25} \cellcolor{white} & \cellcolor{white}\multirow{-2}{*}{2,593} & \cellcolor{white}\multirow{-2}{*}{2.5} &  AMQ & \textbf{45.41} & \textbf{3.49} \\ \cmidrule{2-6}
 & & & BitStack & 44.87 & 10.70 \\
 \rowcolor{gray!25} \cellcolor{white} & \cellcolor{white}\multirow{-2}{*}{3,009} & \cellcolor{white}\multirow{-2}{*}{3} &  AMQ & \textbf{53.57} & \textbf{16.00} \\ \cmidrule{2-6}
 & & & BitStack & 52.70 & 17.29 \\
 \rowcolor{gray!25} \cellcolor{white} & \cellcolor{white}\multirow{-2}{*}{3,425} & \cellcolor{white}\multirow{-2}{*}{3.5}&  AMQ & \textbf{59.12} & \textbf{27.22} \\ \cmidrule{2-6}
  & & & BitStack & 54.20 & 22.44 \\
 \rowcolor{gray!25} \cellcolor{white}\multirow{-12}{*}{\textbf{\large Mistral 7B v0.3}} & \cellcolor{white}\multirow{-2}{*}{3,841} & \cellcolor{white}\multirow{-2}{*}{4} &  AMQ & \textbf{61.12} & \textbf{29.64} \\ \midrule
& 15,317 & 16 & FP16 & 65.20 & 50.57 \\ \cmidrule{2-6}
 &  & & BitStack & 30.63 & 1.59 \\
 \rowcolor{gray!25} \cellcolor{white} & \cellcolor{white}\multirow{-2}{*}{4,085} & \cellcolor{white}\multirow{-2}{*}{2.5} &  AMQ & \textbf{32.10} & \textbf{2.35} \\ \cmidrule{2-6}
 & & & BitStack & 44.03 & 4.55 \\
 \rowcolor{gray!25} \cellcolor{white} & \cellcolor{white}\multirow{-2}{*}{4,501} & \cellcolor{white}\multirow{-2}{*}{3} &  AMQ & \textbf{52.78} & \textbf{19.86} \\ \cmidrule{2-6}
 & & & BitStack & 52.35 & 11.68 \\
 \rowcolor{gray!25} \cellcolor{white} & \cellcolor{white}\multirow{-2}{*}{4,917} & \cellcolor{white}\multirow{-2}{*}{3.5} &  AMQ & \textbf{60.68} & \textbf{31.92} \\ \cmidrule{2-6}
 & & & BitStack & 57.18 & 20.85 \\
 \rowcolor{gray!25} \cellcolor{white}\multirow{-11}{*}{\textbf{\large Llama 3.1 8B}} & \cellcolor{white}\multirow{-2}{*}{5,333} & \cellcolor{white}\multirow{-2}{*}{4} &  AMQ & \textbf{62.68} & \textbf{40.41} \\ \midrule
& 28,171 & 16 & FP16 & 79.85 & 87.72 \\ \cmidrule{2-6}
 & & & BitStack & 60.24 & 34.95 \\
 \rowcolor{gray!25} \cellcolor{white} & \cellcolor{white}\multirow{-2}{*}{6,909} & \cellcolor{white}\multirow{-2}{*}{2.5} &  AMQ & \textbf{63.65} & \textbf{36.69} \\ \cmidrule{2-6}
 & & & BitStack & 67.66 & 65.73 \\
 \rowcolor{gray!25} \cellcolor{white} & \cellcolor{white}\multirow{-2}{*}{7,697} & \cellcolor{white}\multirow{-2}{*}{3} &  AMQ & \textbf{71.61} & \textbf{73.39} \\  \cmidrule{2-6}
 & & & BitStack & 72.55 & 72.02 \\
 \rowcolor{gray!25} \cellcolor{white} & \cellcolor{white}\multirow{-2}{*}{8,484} & \cellcolor{white}\multirow{-2}{*}{3.5} &  AMQ & \textbf{76.93} & \textbf{80.52} \\  \cmidrule{2-6}
 & & & BitStack & 74.43 & 79.30 \\
 \rowcolor{gray!25} \cellcolor{white}\multirow{-11}{*}{\textbf{\large Qwen2.5 14B}}  & \cellcolor{white}\multirow{-2}{*}{9,272} & \cellcolor{white}\multirow{-2}{*}{4} &  AMQ & \textbf{78.69} & \textbf{83.24} \\ 
 \bottomrule \bottomrule
\end{tabular}
}

\vspace{-0.3cm}
\caption{5-shot MMLU, GSM8K task results over Mistral 7B v0.3, Llama 3.1 8B and Qwen2.5 14B.}
\vspace{-0.6cm}
\label{tab:mmlu_gsm8k}
\end{table}

\renewcommand{\arraystretch}{0.75}
\begin{table}[thb!]
\setlength{\tabcolsep}{1mm}
\resizebox{\columnwidth}{!}
{
\begin{tabular}{c|c|c|c||cc||c}
\midrule
\textbf{\large Model} &  \textbf{\begin{tabular}[c]{@{}c@{}}\textbf{\large{Mem.}}\\ \textbf{\large{(MB)}}\end{tabular}} & \textbf{\begin{tabular}[c]{@{}c@{}}\textbf{\large{Avg.}}\\ \textbf{\large{Bits}}\end{tabular}} &  \textbf{\large Method} &  \textbf{\large Wiki($\downarrow$)} &  \textbf{\large C4($\downarrow$)} & \textbf{\large Avg.($\uparrow$)} \\ \midrule
 & 12,853 & 16 & FP16 & 5.47 & 7.26 & 70.49 \\ \cmidrule{2-7}
 & \multirow{2}{*}{2,238}  & \multirow{2}{*}{$2.25$} 
 & GPTQ$_{w2g128}$ & 61.77 & 44.10 & 43.19 \\
 & & & AWQ$_{w2g128}$ & 2.22e5 & 1.68e5 & 36.12 \\
\rowcolor{gray!25} \cellcolor{white} & \cellcolor{white}2,315 & \cellcolor{white}$2.35$ & AMQ & \textbf{11.49} & \textbf{15.12} & \textbf{57.86} \\ \cmidrule{2-7}
 & \multirow{3}{*}{2,817} & \multirow{3}{*}{$3.0$} 
 & GPTQ$_{w3}$ & 9.27 & 11.81 & 60.70 \\
 & & & AWQ$_{w3}$ & 15.45 & 17.44 & 54.67 \\
\rowcolor{gray!25} \cellcolor{white} & \cellcolor{white} & \cellcolor{white} & AMQ & \textbf{6.83} & \textbf{9.03} & \textbf{65.59} \\ \cmidrule{2-7}
 & \multirow{3}{*}{3,010} & \multirow{3}{*}{$3.25$}
 & GPTQ$_{w3g128}$& 6.45 & 8.53 & 67.22 \\ 
 & & & AWQ$_{w3g128}$ & 6.25 & 8.30 & 67.63 \\
\rowcolor{gray!25} \cellcolor{white} & \cellcolor{white} & \cellcolor{white} & AMQ & \textbf{6.20} & \textbf{8.25} & \textbf{67.94} \\ \cmidrule{2-7}
 & \multirow{3}{*}{3,589} & \multirow{3}{*}{$4.0$} 
 & GPTQ$_{w4}$ & 6.09 & 7.86 & 68.55 \\ 
 & & & AWQ$_{w4}$ & 5.83 & 7.72 & 69.10 \\
\rowcolor{gray!25} \cellcolor{white}\multirow{-16}{*}{\textbf{\large 7B}} & \cellcolor{white} & \cellcolor{white} & AMQ & \textbf{5.68} & \textbf{7.54} & \textbf{69.29} \\ \midrule
 & 24,826 & 16 & FP16 & 4.88 & 6.73 & 73.23 \\ \cmidrule{2-7}
 & \multirow{2}{*}{4,029}& \multirow{2}{*}{$2.25$} 
 & GPTQ$_{w2g128}$ & 27.78 & 23.39 & 45.64 \\
 & & & AWQ$_{w2g128}$ & 1.22e5 & 9.55e4 & 40.59 \\
\rowcolor{gray!25} \cellcolor{white} & \cellcolor{white}4,181 & \cellcolor{white}$2.35$ & AMQ & \textbf{7.60} & \textbf{10.29} & \textbf{64.59} \\ \cmidrule{2-7}
 & \multirow{3}{*}{5,164} & \multirow{3}{*}{$3.0$} 
 & GPTQ$_{w3}$ & 6.75 & 8.96 & 67.39 \\
 & & & AWQ$_{w3}$ & 6.45 & 9.07 & 67.34 \\
\rowcolor{gray!25} \cellcolor{white} & \cellcolor{white} & \cellcolor{white} & AMQ & \textbf{5.68} & \textbf{7.80} & \textbf{70.31} \\ \cmidrule{2-7}
 & \multirow{3}{*}{5,542} & \multirow{3}{*}{$3.25$} 
 & GPTQ$_{w3g128}$ & 5.48 & 7.49 & 70.89 \\
 & & & AWQ$_{w3g128}$ & \textbf{5.32} & \textbf{7.31} & \textbf{72.11} \\
\rowcolor{gray!25} \cellcolor{white} & \cellcolor{white} & \cellcolor{white} & AMQ & 5.36 & 7.33 & 71.52 \\ \cmidrule{2-7}
 & \multirow{3}{*}{6,676} & \multirow{3}{*}{$4.0$} 
 & GPTQ$_{w4}$ & 5.19 & 7.06 & 71.85 \\
 & & & AWQ$_{w4}$ & 5.06 & 6.96 & 72.59 \\
\rowcolor{gray!25} \cellcolor{white}\multirow{-16}{*}{\textbf{\large 13B}} & \cellcolor{white} & \cellcolor{white} & AMQ & \textbf{5.03} & \textbf{6.91} & \textbf{72.98} \\ \midrule
& 131,563 & 16 & FP16 & 3.32 & 5.71 & 77.75 \\ \cmidrule{2-7}
 & \multirow{2}{*}{19,363} & \multirow{2}{*}{$2.25$} 
 & GPTQ$_{w2g128}$ & 8.33 & 10.71 & 59.85 \\
 & & & AWQ$_{w2g128}$ & 7.25e4 & 6.56e4 & 40.54 \\
 \rowcolor{gray!25} \cellcolor{white} & \cellcolor{white}20,179 & \cellcolor{white}$2.35$ &  AMQ & \textbf{5.17} & \textbf{7.59} & \textbf{73.45} \\ \cmidrule{2-7}
 & \multirow{3}{*}{25,483} & \multirow{3}{*}{$3.0$} 
 & GPTQ$_{w3}$ & 4.88 & 7.11 & 73.31 \\
 & & & AWQ$_{w3}$ & 4.36 & 6.63 & 75.10 \\
\rowcolor{gray!25} \cellcolor{white} & \cellcolor{white}& \cellcolor{white} & AMQ & \textbf{4.01} & \textbf{6.29} & \textbf{76.05} \\ \cmidrule{2-7}
 & \multirow{3}{*}{27,523} & \multirow{3}{*}{$3.25$} 
 & GPTQ$_{w3g128}$ & 3.88 & 6.11 & \textbf{76.64} \\
 & & & AWQ$_{w3g128}$ & 3.74 & 6.04 & 76.58 \\
 \rowcolor{gray!25} \cellcolor{white} & \cellcolor{white} & \cellcolor{white} & AMQ & \textbf{3.73} & \textbf{6.03} & 76.38 \\ \cmidrule{2-7}
 & \multirow{3}{*}{33,643} & \multirow{3}{*}{$4.0$} 
 & GPTQ$_{w4}$ & 3.59 & 5.90 & 77.07 \\
 & & & AWQ$_{w4}$ & 3.48 & 5.84 & 77.41 \\
\rowcolor{gray!25} \cellcolor{white}\multirow{-16}{*}{\textbf{\large 70B}} & \cellcolor{white} & \cellcolor{white} & AMQ & \textbf{3.46} & \textbf{5.80} & \textbf{77.48} \\ \bottomrule \bottomrule
\end{tabular}
}
\vspace{-0.3cm}
\caption{Evaluation of Llama 2 7B/13B/70B models quantized by AMQ, AWQ, and GPTQ on WikiText-2, C4 perplexity (PPL), and zero-shot tasks. For 2.25-bit settings, our method matches asymmetric clipping in AWQ; thus, we report results with an additional 0.1 bits. Memory overhead from extra quantization parameters in GPTQ and AWQ at w3, w4 is omitted as it is negligible. Detailed zero-shot accuracy is provided in \Cref{tab:Llama-2-fixed-precision-quantization}.}
\vspace{-0.1cm}

\vspace{-0.6cm}
\label{tab:table_3}
\end{table}


To validate the superiority of AMQ, we conducted a series of analyses and experiments using the Llama 2 model family \cite{llama2}, ranging in size from 7B to 70B. Also, we further provide the experimental results on Llama 3.1 8B, 70B \cite{llama3}, Qwen2.5-7B, 14B \cite{qwen2.5} and Mistral 7B v0.3 \cite{mistral7b} in Appendix~\ref{additional_expr}. We employed 128 samples from the WikiText-2 train set \cite{wikitext2} as the calibration set to measure the sensitivity and evaluate the models identified during the search. 

To assess the effectiveness of the bit configurations discovered by AMQ, we utilized AWQ \cite{awq}, a well-established weight-only quantization method, following its original settings except for asymmetric clipping \cite{LLMC}. AMQ utilized a group size of 128 for search and performance evaluation.
We further compared AMQ with PB-LLM \cite{PB-LLM} and BitStack \cite{bitstack} under varying memory constraints, and also with fixed-precision quantization methods, GPTQ and AWQ.

We evaluate the optimized models by reporting language-modeling perplexity (PPL) on the WikiText-2 test set and the C4 validation set \cite{c4}. For zero-shot performance, we use the LM Evaluation Harness \cite{lm_eval} on six benchmarks: ARC-Challenge, ARC-Easy \cite{arc}, PIQA \cite{piqa}, HellaSwag \cite{hellaswag}, BoolQ \cite{boolq}, and WinoGrande \cite{winogrande}. To probe more demanding tasks, we further evaluate 5-shot MMLU \cite{mmlu} and GSM8K \cite{gsm8k}. We also report inference throughput as the median tokens per second when generating 128 tokens with batch size 1. Additional details are provided in Appendix~\ref{detailed_expr_setting}.



\subsection{AMQ vs. Any-Size Compression}
\Cref{tab:main_table} shows the perplexity and zero-shot task performance of the optimized models under varying memory budgets and models using AMQ, BitStack and PB-LLM. AMQ consistently outperforms BitStack across multiple bit-width settings and model scales. At extremely low precision (e.g., 2.5 bits), AMQ achieves the best average zero-shot accuracy among all methods. Notably, for the 70B model at 3.5 bits, AMQ retains up to 99.18\% of the FP16 model's average zero-shot performance. 
Even with 0.5 fewer bits, AMQ matches or exceeds the performance of the  3.5 bits baselines over different model scales, demonstrating its robustness.
As shown in \Cref{fig:figure_1}, AMQ consistently outperforms baselines across all bit levels.
Moreover, \Cref{tab:mmlu_gsm8k} shows that AMQ consistently outperforms BitStack on 5-shot MMLU and GSM8K across models, demonstrating superior performance on challenging tasks. This highlights the efficiency and generalization capability of our search strategy.

\subsection{Inference Acceleration}


As illustrated in \Cref{fig:Inference_speed}, weight decomposition-based compression methods (BitStack) suffer from the overhead of weight reconstruction during inference.
In contrast, AMQ customizes kernels for each linear layer based on its bit configuration, resulting in up to 2.67× speedup compared to FP16 on the L40S GPU for Llama 2 7B. For Llama 2 13B, AMQ achieves an even higher speedup of 3.16×. Moreover, AMQ delivers high-speed inference while maintaining a small memory footprint, making it particularly effective in memory-constrained environments such as the RTX 3090.

\subsection{AMQ vs. Fixed-Precision Quantization}

We compared the performance of AMQ with existing quantization methods, GPTQ and AWQ, which using iso-precision over all layers. \Cref{tab:table_3} presents the perplexity and average zero-shot accuracy across bit-widths ranging from 2.25 to 4 bits. AMQ consistently matches or surpasses uniform quantization, validating the effectiveness of its discovered bit configurations. For Llama 2 7B, AMQ maintains stable accuracy at 2.35 bits, while GPTQ and AWQ degrade sharply at 2.25 bits. It remains robust at 3 bits and competitive even at 4 bits across model sizes. On Llama 2 13B, AMQ achieves 96.01\% of FP16 average zero-shot accuracy with only 3 bits, demonstrating strong memory efficiency without sacrificing performance.

\section{Analysis}
\begin{figure}[tb!]
\centering
\includegraphics[width=\columnwidth]{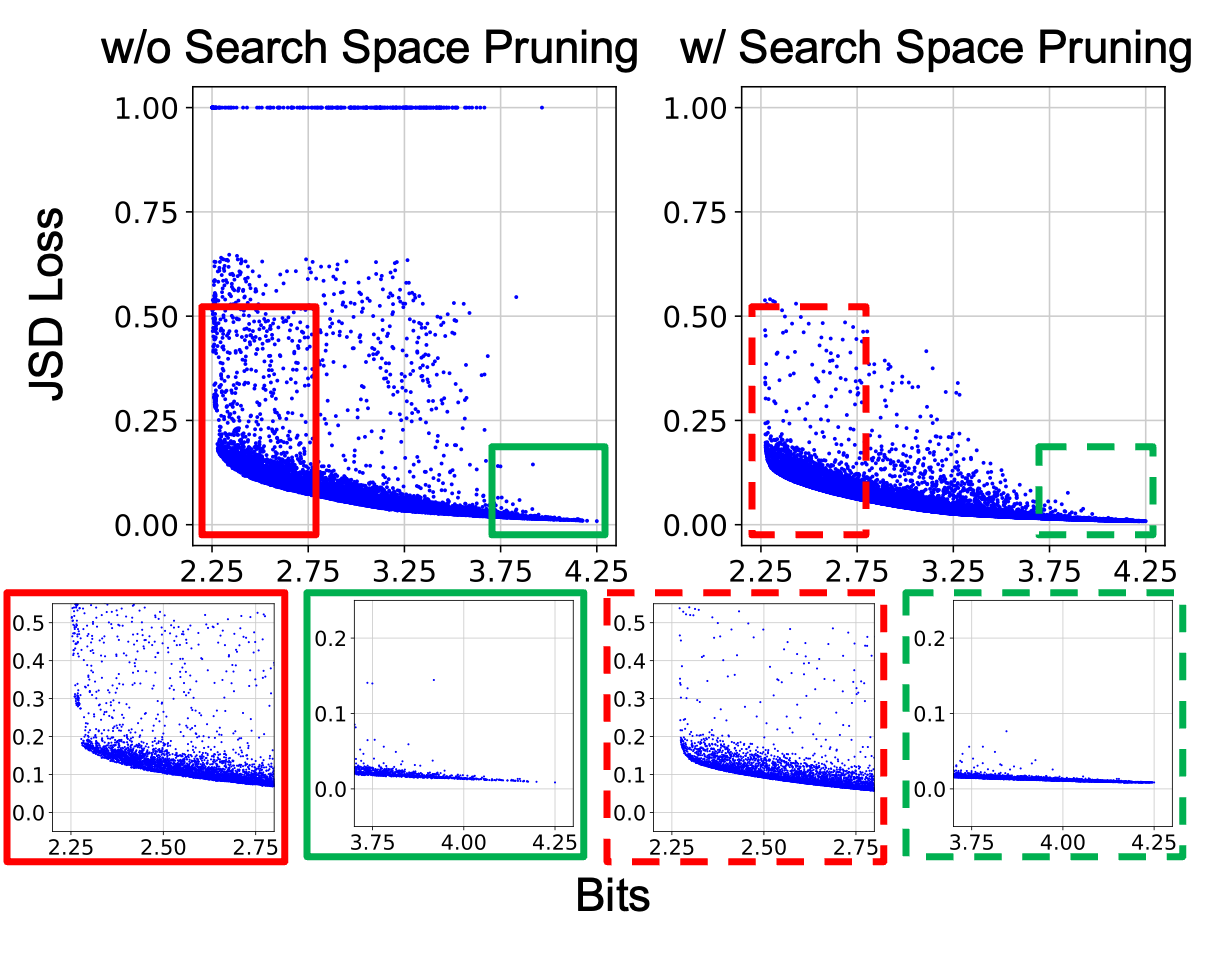}
\vspace{-1cm}
\caption{Total search samples on Llama 2 70B with vs. without Search Space Pruning.}
\vspace{-0.4cm}
\label{fig:outlier_archive}
\end{figure}

\begin{figure}[tb!]
\centering
\includegraphics[width=\columnwidth]{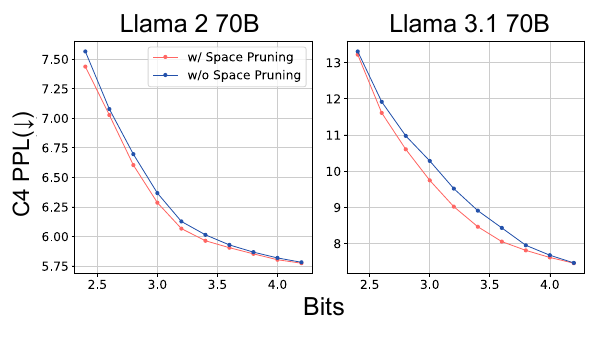}
\vspace{-1cm}
\caption{C4 perplexity of Llama 2 and 3.1 70B, with and without search space pruning.}
\vspace{-0.5cm}
\label{fig:outlier_perplexity}
\end{figure}


\renewcommand{\arraystretch}{0.6}
\begin{table*}[ht!]
\resizebox{\textwidth}{!}{
\begin{tabular}{c||c|c||c|c||c|c||c|c||c|c||c|c}
\midrule
\textbf{\large Model} & \multicolumn{4}{c||}{\textbf{\large 7B}} & \multicolumn{4}{c||}{\textbf{\large 13B}} & \multicolumn{4}{c}{\textbf{\large 70B}} \\ \midrule
Type & \multicolumn{2}{c||}{Search} & \multicolumn{2}{c||}{Compression} & \multicolumn{2}{c||}{Search} & \multicolumn{2}{c||}{Compression} & \multicolumn{2}{c||}{Search} & \multicolumn{2}{c}{Compression} \\ \midrule
Parameter & \#GPU & Cost (h) & \#GPU & Cost (h) & \#GPU & Cost (h) & \#GPU & Cost (h) & \#GPU & Cost (h) & \#GPU & Cost (h) \\ \midrule
AWQ & - & - & 1 & 0.15 & - & - & 1 & 0.28 & - & - & 1 & 1.5 \\
OmniQuant & - & - & 1 & 1.24 & - & - & 1 & 2.2 & - & - & 1 & 10 \\
BitStack & 1 & 10 & 1 & 2.25 & 1 & \textgreater{}72 & 1 & 4.25 & 2 & \textgreater{}300 & 1 & 23 \\
AMQ & 1 & 5 & 1 & 0.15 & 1 & 8 & 1 & 0.28 & 2 & 44 & 1 & 1.5 \\ \bottomrule \bottomrule
\end{tabular}
}
\vspace{-0.2cm}
\caption{The search and compression time on Llama 2 family of AWQ, OmniQuant, BitStack, AMQ.}
\vspace{-0.6cm}
\label{tab:search_compression_time}
\end{table*}
\subsection{Search and Compression Cost}

We compare the overall algorithmic costs of AWQ, OmniQuant \cite{omniquant}, and BitStack with AMQ in \Cref{tab:search_compression_time}. The comparison includes both search time, required for exploring memory-performance trade-offs, and compression time, needed to generate optimized models, using the Llama 2 family on NVIDIA A100-80GB GPUs. AWQ and OmniQuant incur no search overhead but are constrained to fixed-bit quantization, limiting flexibility. OmniQuant also requires fine-tuning, increasing its compression cost. BitStack performs only one compression run by weight decomposition but suffers from significantly higher search and compression time due to block evaluation and sorting. AMQ achieves practical search costs by avoiding type conversion overhead through a quantization proxy and accelerating convergence via a quality predictor and search space pruning. For instance, on Llama 2 7B, AMQ evaluates only 10,250 candidates directly, while predicting performance for over 800,000 samples, despite the vast search space size (approx.\ $10^{100}$).

\subsection{Effect of Search Space Pruning}\label{sec:prune}

\textbf{Impact on Vast Space.}
We assess the effect of pruning on large search space by comparing results with and without it. As shown in \Cref{fig:outlier_perplexity}, pruning markedly improves search quality for Llama 2 70B and Llama 3.1 70B.
Without pruning, the search fails to explore configurations near 4.25 bits at Llama 2 70B, as those regions remain entirely unvisited, illustrated in \Cref{fig:outlier_archive}. This indicates that outlier linear layers destabilize training and inject noise into quality predictions, steering the search toward suboptimal points, especially for large scale models.

\textbf{Ablation on Threshold and Calibration Set.}
We investigate how the selection of outlier layers is affected by the calibration set and threshold. Note that we set the sensitivity threshold conservatively to prevent excluding too many candidates and limiting the expressiveness of the search space.
\Cref{tab:threshold_sensitivity} shows that sensitive layers consistently occur in early V linear layers of self-attention and early/late Down linear layers of MLP, regardless of calibration set. The excluded fraction remains small (0.45–2.14\%) with negligible impact on C4 perplexity, and AMQ converges stably as long as the threshold is not overly strict. We thus adopt twice the median threshold, which yields consistently robust performance. 

\subsection{Robustness over Random Seed}
\begin{figure}[t]
    \centering
    \includegraphics[width=0.8\columnwidth]{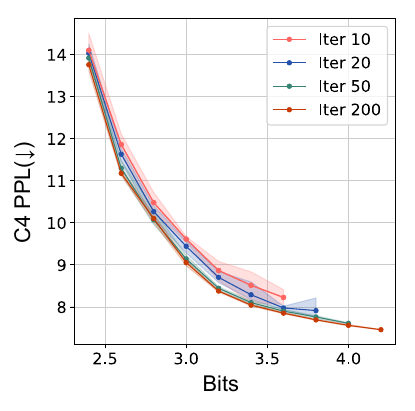}
    \vspace{-0.4cm}
    \caption{C4 perplexity variation of Pareto frontiers across iterations on Llama 2 7B with six random seeds. Data points are plotted only when all seeds discover a sample for the corresponding bit-width at a given iteration.}
    \vspace{-0.6cm}
    \label{fig:random_seed}
\end{figure}

\renewcommand{\arraystretch}{0.75}
\begin{table*}[t]
\resizebox{\textwidth}{!}{%
\begin{tabular}{c|c|c|c|c||c|c|c|c}
\midrule
\textbf{\large Model} & \textbf{\large Dataset} & \textbf{\begin{tabular}[c]{@{}c@{}}\large Threshold\\ \large (×median)\end{tabular}} & \textbf{\begin{tabular}[c]{@{}c@{}}\large Outlier\\ \large Layer\end{tabular}} & \textbf{\begin{tabular}[c]{@{}c@{}}\large \# Outlier\\ \large Layer\end{tabular}} & \textbf{\large 2.5-bit} & \textbf{\large 3-bit} & \textbf{\large 3.5-bit} & \textbf{\large 4-bit} \\ \midrule
\multirow{10}{*}{\textbf{\large Mistral 7B v0.3}} & WikiText-2 & \multirow{2}{*}{1.5} & \multirow{2}{*}{V: {[}3{]}, Down: {[}1, 31{]}} & \multirow{2}{*}{3 (1.34\%)} & \multirow{2}{*}{\textbf{12.61}} & \multirow{2}{*}{10.05} & \multirow{2}{*}{9.04} & \multirow{2}{*}{\textbf{8.73}} \\
 & C4 &  &  &  &  &  &  &  \\ \cmidrule{2-9}
 & WikiText-2 & \multirow{2.7}{*}{2} & Down: {[}1{]} & 1 (0.45\%) & 12.67 & 10.08 & \textbf{9.03} & 8.74 \\ \cmidrule{2-2} \cmidrule{4-9}
 & C4 &  & Down: {[}1, 31{]} & 2 (0.89\%) & 12.87 & \textbf{9.97} & \textbf{9.03} & \textbf{8.73} \\ \cmidrule{2-9}
 & WikiText-2 & \multirow{2}{*}{3} & \multirow{4}{*}{Down: {[}1{]}} & \multirow{4}{*}{1 (0.45\%)} & \multirow{4}{*}{12.67} & \multirow{4}{*}{10.08} & \multirow{4}{*}{\textbf{9.03}} & \multirow{4}{*}{8.74} \\
 & C4 &  &  &  &  &  &  &  \\ \cline{2-2} \noalign{\vskip 2pt}
 & WikiText-2 & \multirow{2}{*}{5} &  &  &  &  &  & \\
 & C4 &  &  &  &  &  &  &  \\ \midrule
\multirow{11}{*}{\textbf{\large Llama 2 13B}} & WikiText-2 & \multirow{2}{*}{1.5} & \multirow{2}{*}{V: {[}0, 1, 2{]}, Down: {[}0, 3,   39{]}} & \multirow{2}{*}{6 (2.14\%)} & \multirow{2}{*}{9.43} & \multirow{2}{*}{7.78} & \multirow{2}{*}{\textbf{7.13}} & \multirow{2}{*}{\textbf{6.91}} \\ 
 & C4 &  &  &  &  &  &  &  \\ \cmidrule{2-9}
 & WikiText-2 & \multirow{2.7}{*}{2} & V: {[}0{]}, Down: {[}0, 3{]} & 3 (1.07\%) & 9.39 & 7.77 & \textbf{7.13} & \textbf{6.91} \\ \cmidrule{2-2} \cmidrule{4-9}
 & C4 &  & V: {[}0, 1{]}, Down: {[}0,   3{]} & 4 (1.43\%) & \textbf{9.38} & \textbf{7.75} & \textbf{7.13} & \textbf{6.91} \\ \cmidrule{2-9}
 & WikiText-2 & \multirow{2}{*}{3} & \multirow{2}{*}{V: {[}0{]}, Down: {[}0, 3{]}} & \multirow{2}{*}{3 (1.07\%)} & \multirow{2}{*}{9.39} & \multirow{2}{*}{7.77} & \multirow{2}{*}{\textbf{7.13}} & \multirow{2}{*}{\textbf{6.91}} \\
 & C4 &  &  &  &  &  &  &  \\ \cmidrule{2-9}
 & WikiText-2 & \multirow{2.7}{*}{5} & Down: {[}0, 3{]} & 2 (0.71\%) & 9.45 & 7.78 & 7.15 & \textbf{6.91} \\ \cmidrule{2-2} \cmidrule{4-9}
 & C4 &  & V: {[}0{]}, Down: {[}0, 3{]} & 3 (1.07\%) & 9.46 & 7.80 & \textbf{7.13} & \textbf{6.91} \\ \bottomrule \bottomrule
\end{tabular}
}
\vspace{-0.3cm}
\caption{C4 Perplexity and selected outlier linear layers over different calibration sets and sensitivity thresholds for Search Space Pruning. The default is WikiText-2 and $2 \times \text{median}$. Layer indices start at 0. 
We employ 32 samples from the C4 calibration set to approximate the token count of 128 samples from WikiText-2.
}
\vspace{-0.6cm}
\label{tab:threshold_sensitivity}
\end{table*}

\begin{figure}[t]
\centering
\includegraphics[width=\columnwidth]{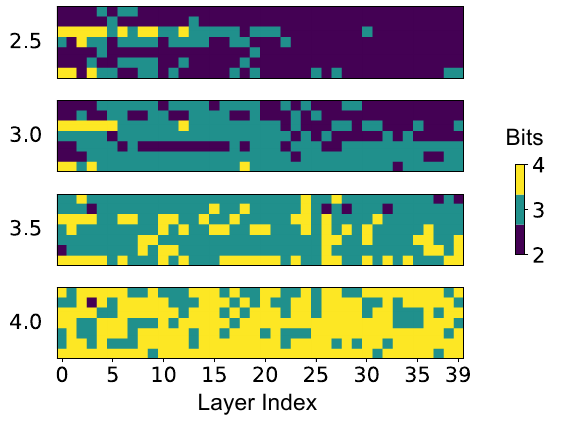}
\vspace{-0.8cm}
\caption{Visualization of bit allocation over linear layers with different average bits at Llama 2 13B. Each row in each box represents Q, K, V, O, Gate, Up, and Down. The numbers on the left indicate the average bits per configuration.}
\vspace{-0.6cm}
\label{fig:bit_assign_13b}
\end{figure}

\Cref{fig:random_seed} illustrates the perplexity variations of the Pareto frontiers on the C4 validation set across different bit-widths at iterations 10, 20, 50, and 200 (final) during AMQ search on the Llama 2 7B model with six random seeds. The results highlight the robustness of the search process, which consistently converges to an optimal Pareto frontier while progressively reducing variation at each iteration, regardless of the initial random seed.



\subsection{Visualization of Bit Allocation}

Figure \ref{fig:bit_assign_13b} shows bit allocation across linear layers for models quantized to average bit-widths of 2.5, 3.0, 3.5, and 4.0. As bit-width decreases, Query and Key layers in self-attention are prioritized for lower bit-widths, followed by the Gate layer in the MLP. Notably, the Value layer in self-attention consistently retains a higher bit-width, underscoring its critical role in preserving model performance.




\section{Conclusion}


In this paper, we propose AMQ, an automated mixed-precision weight-only quantization framework designed to achieve optimal model quality under memory constraints. Our approach precisely defines the search space, and by selectively excluding low-bit-sensitive outlier layers, we effectively prune the initial search space, enhancing both convergence speed and search quality. Additionally, we leverage a quantization proxy to generate quantized models rapidly. Finally, we introduce a quality predictor that estimates the performance of unseen bit-width combinations, significantly reducing the evaluation overhead during the search process. Our experimental results demonstrate that AMQ efficiently allocates bit-widths to each linear layer, even at lower precision levels, outperforming existing baselines while effectively addressing real-world constraints.

\section*{Limitations}
This study presents an efficient approach to exploring the search space using a quality score predictor, a genetic algorithm, and search space shrinking to address the NP-complete problem of optimal bit configuration in memory-constrained environments. However, alternative solvers for NP-complete problems may yield better configurations, which will be investigated in future work.

The current method focuses on weight-only quantization, addressing primary memory constraints in LLMs. Future research will extend this to tackle computation-bound challenges through activation quantization and explore rotation-based techniques to further improve performance and efficiency.

\newpage

\section*{Acknowledgements}
This work was supported by IITP and NRF grant funded by the Korea government(MSIT) (No. RS-2023-00213611, RS-2024-00457882, RS-2024-00396013) and Samsung Advanced Institute of Technology.

\bibliography{acl_latex}

\clearpage
\appendix
\section{Proof of Motivation in \Cref{thm:pareto_frontier_preservation}}
\label{proof_pareto_frontier}
Since $Q_1$ and $Q_2$ are injective, they induce strict total orders on $X$. The condition implies order equivalence for all $x, y \in X$: 
\[
Q_1(x)<Q_1(y)\;\iff\; Q_2(x)<Q_2(y).
\]
Let $\mathcal{F}_1$ and $\mathcal{F}_2$ denote the Pareto-frontiers for $(Q_1,S)$ and $(Q_2,S)$, respectively. Suppose $a\in\mathcal{F}_1$ but $a\notin\mathcal{F}_2$. Then there exists $b\in X$ such that $Q_2(b)>Q_2(a)$ and $S(b)\le S(a)$, which by order equivalence implies $Q_1(b)>Q_1(a)$, contradicting $a\in\mathcal{F}_1$. Hence $\mathcal{F}_1\subseteq\mathcal{F}_2$. The same argument with $Q_1$ and $Q_2$ swapped yields $\mathcal{F}_2\subseteq\mathcal{F}_1$. Therefore, $\mathcal{F}_1=\mathcal{F}_2$.



\section{Detailed Algorithm}
\begin{algorithm}[ht]
\caption{Auto Weight Quantization Search}
\label{alg:search}
\begin{algorithmic}[1]
\Require {
$\mathcal{S}$: Search space,
$Q_2, Q_3, Q_4$: Proxy models for 2-bit, 3-bit, and 4-bit quantization,
$\mathcal{D}$: Calibration dataset,
$N$: \# of initial samples, $I$: \# of iterations, B: Target bits
}

\State $\mathcal{S} \gets \text{SpaceShrink}(\mathcal{S}, \mathcal{D})$  
\State $\mathcal{A} \gets \emptyset$ \Comment{Initialize archive}

\For{$i = 1$ to $N$}  \Comment{Initial sampling}
    \State $\alpha \gets \text{RandomSample}(\mathcal{S})$
    \State $Q_{\alpha} \gets \text{Assemble}(\alpha, Q_2, Q_3, Q_4)$
    \State $\text{Score} \gets \text{Evaluate}(Q_{\alpha}, \mathcal{D})$
    \State $\mathcal{A} \gets \mathcal{A} \cup (\alpha, \text{Score})$
\EndFor

\For{$j = 1$ to $I$} \Comment{Iterative search}
    \State $\mathcal{P} \gets \text{Re/TrainPredictor}(\mathcal{A})$
    \State $\text{front} \gets \text{ParetoSort}(\mathcal{A})$
    \State $\bar{\alpha} \gets \text{NSGA-II}(\text{front}, \mathcal{P})$ 

    \For{$\alpha \in \bar{\alpha}$} \Comment{Verify candidates}
        \State $Q_{\alpha} \gets \text{Assemble}(\alpha, Q_2, Q_3, Q_4)$
        \State $\text{Score} \gets \text{Evaluate}(Q_{\alpha}, \mathcal{D})$
        \State $\mathcal{A} \gets \mathcal{A} \cup (\alpha, \text{Score})$
    \EndFor
\EndFor

\State $\alpha^* \gets \text{SelectOptimal}(\mathcal{A}, B)$
\State \Return $\alpha^*$

\end{algorithmic}
\end{algorithm}

\section{Detailed Experimental Setting}
\label{detailed_expr_setting}
All experiments were run on up to two NVIDIA A100 80 GB GPUs. For AMQ, for each target bit precision we evaluated the candidate whose average bit-width lay within ±0.005 of the target and achieved the lowest score. During Search Space Pruning, we quantified each linear layer’s sensitivity using the JSD score and the same calibration set used for model-quality evaluation within Iterative search-and-update process. For PB-LLM, we set the group size to 128, counting only weight memory and excluding any additional indexing overhead. For BitStack, we used the official pre-trained weights from their implementation.

In all inference speed experiments including FP16, QuIP\#, and BitStack, we leveraged the multi-head attention and layer normalization kernels with FasterTransformer \cite{FasterTransformer}. 
For the 4-bit linear layers, AMQ employed TensorRT-LLM \cite{TensorRT-LLM}-based kernels, while for the 2-bit and 3-bit linear layers, AMQ utilized AutoGPTQ \cite{AutoGPTQ} kernels.

The hyperparameters used for searching the Llama 2 family in our experiments are detailed in \Cref{table:hyper_parameter}.


\renewcommand{\arraystretch}{0.75}
\begin{table}[ht]
\centering
\begin{tabular}{c|ccc}
\midrule
\multirow{2}{*}{\textbf{Hyper-parameter}} & \multicolumn{3}{c}{\textbf{Model}} \\ \cmidrule{2-4}
 & \textbf{7B} & \textbf{13B} & \textbf{70B} \\ \midrule
Search Iteration & 200 & 200 & 250 \\
NSGA-II Candidate & 50 & 50 & 50 \\
Pretraining Data & 250 & 300 & 600 \\
NSGA-II Pop & 200 & 200 & 200 \\
NSGA-II Iteration & 20 & 20 & 20 \\
Cross-over Probability & 0.9 & 0.9 & 0.9 \\
Mutation Probability & 0.1 & 0.1 & 0.1 \\
Subset Pop & 100 & 100 & 100 \\ \bottomrule \bottomrule
\end{tabular}
\vspace{-0.3cm}
\caption{Hyper-parameters of our algorithm used in Llama 2 search space.}
\vspace{-0.6cm}
\label{table:hyper_parameter}
\end{table}

\section{Robustness over NSGA-II Hyperparameter}

\renewcommand{\arraystretch}{0.75}
\begin{table}[h]
\resizebox{\columnwidth}{!}{
\begin{tabular}{c|c||cc}
\midrule
\begin{tabular}[c]{@{}c@{}}\large Average\\\large Bits\end{tabular} & \begin{tabular}[c]{@{}c@{}}\large Crossover\\\large Prob.\end{tabular} & \large Wiki(↓) & \large C4(↓) \\ \midrule
\multirow{3}{*}{2.5} & 0.5 & 9.26 & \textbf{12.32} \\
 & 0.7 & \textbf{9.19} & \textbf{12.32} \\
 & 0.9 & 9.24 & 12.37 \\ \midrule
\multirow{3}{*}{3} & 0.5 & \textbf{6.78} & \textbf{9.03} \\
 & 0.7 & 6.84 & 9.07 \\
 & 0.9 & 6.83 & \textbf{9.03} \\ \midrule
\multirow{3}{*}{3.5} & 0.5 & 5.93 & 7.89 \\
 & 0.7 & \textbf{5.92} & \textbf{7.88} \\
 & 0.9 & 5.95 & 7.90 \\ \midrule
\multirow{3}{*}{4} & 0.5 & \textbf{5.67} & \textbf{7.54} \\
 & 0.7 & 5.68 & \textbf{7.54} \\
 & 0.9 & 5.68 & \textbf{7.54} \\ \bottomrule \bottomrule
\end{tabular}
}
\vspace{-0.3cm}
\caption{Evaluation of different crossover probabilities over Llama 2 7B. Our default option is 0.9.}
\vspace{-0.4cm}
\label{tab:crossover_expr}
\end{table}

\renewcommand{\arraystretch}{0.75}
\begin{table}[ht]
\resizebox{\columnwidth}{!}{
\begin{tabular}{c|c||cc}
\midrule
\begin{tabular}[c]{@{}c@{}}\large Average\\\large Bits\end{tabular} & \begin{tabular}[c]{@{}c@{}}\large Crossover\\\large Prob.\end{tabular} & \large Wiki(↓) & \large C4(↓) \\ \midrule
\multirow{5}{*}{2.5} & 0.01 & \textbf{9.18} & 12.23 \\
 & 0.05 & 9.25 & \textbf{12.22} \\
 & 0.1 & 9.24 & 12.37 \\
 & 0.2 & 9.23 & 12.26 \\
 & 0.3 & 9.26 & 12.31 \\ \midrule
\multirow{5}{*}{3} & 0.01 & 6.90 & 9.07 \\
 & 0.05 & \textbf{6.80} & 9.03 \\
 & 0.1 & 6.83 & 9.03 \\
 & 0.2 & 6.84 & \textbf{8.98} \\
 & 0.3 & 6.83 & 9.06 \\ \midrule
\multirow{5}{*}{3.5} & 0.01 & \textbf{5.93} & \textbf{7.88} \\
 & 0.05 & 5.98 & 7.91 \\
 & 0.1 & 5.95 & 7.90 \\
 & 0.2 & 5.94 & 7.89 \\
 & 0.3 & 5.95 & 7.90 \\ \midrule
\multirow{5.2}{*}{4} & 0.01 & 5.68 & \textbf{7.54} \\
 & 0.05 & 5.70 & 7.56 \\
 & 0.1 & 5.68 & \textbf{7.54} \\
 & 0.2 & 5.69 & \textbf{7.54} \\
 & 0.3 & \textbf{5.67} & \textbf{7.54} \\ \bottomrule \bottomrule
\end{tabular}
}
\vspace{-0.3cm}
\caption{Evaluation of different mutation probabilities over Llama 2 7B. Our default option is 0.1.}
\vspace{-0.6cm}
\label{tab:mutation_expr}
\end{table}

We assess the robustness of the search process with respect to variations in NSGA-II hyperparameters, specifically the crossover and mutation probabilities. \Cref{tab:crossover_expr} and \Cref{tab:mutation_expr} present the results for the Llama 2 7B model under different settings. The results demonstrate that AMQ consistently maintains strong performance across a wide range of NSGA-II configurations, highlighting the robustness of the method. As no single setting shows clear superiority, we arbitrarily select one for our experiments.

\section{Choice of Search Method, Quantization Proxy, Predictor, and Iteration}

\renewcommand{\arraystretch}{0.75}
\begin{table}[ht]
\resizebox{\columnwidth}{!}{
\begin{tabular}{c|c|c||cc}
\midrule
\begin{tabular}[c]{@{}c@{}}\large Memory\\\large (MB)\end{tabular} & \begin{tabular}[c]{@{}c@{}}\large Avg.\\ \large Bits\end{tabular} & \large Predictor & \large Wiki(↓) & \large C4(↓) \\ \midrule
\multirow{2}{*}{2,431} & \multirow{2}{*}{2.5} & MLP & \textbf{9.24} & \textbf{12.24} \\
 &  & RBF & \textbf{9.24} & 12.37 \\ \midrule
\multirow{2}{*}{2,817} & \multirow{2}{*}{3} & MLP & \textbf{6.83} & 9.07 \\
 &  & RBF & \textbf{6.83} & \textbf{9.03} \\ \midrule
\multirow{2}{*}{3,203} & \multirow{2}{*}{3.5} & MLP & \textbf{5.93} & \textbf{7.89} \\
 &  & RBF & 5.95 & 7.90 \\ \midrule
\multirow{2}{*}{3,589} & \multirow{2}{*}{4} & MLP & \textbf{5.68} & 7.55 \\
 &  & RBF & \textbf{5.68} & \textbf{7.54} \\ \bottomrule \bottomrule
\end{tabular}
}
\vspace{-0.3cm}
\caption{Evaluation of MLP/RBF predictor over Llama 2 7B.}
\vspace{-0.3cm}
\label{tab:mlp_rbf_predictor}
\end{table}

\renewcommand{\arraystretch}{0.75}
\begin{table}[]
\resizebox{\columnwidth}{!}{%
\begin{tabular}{c|c||c||c|c|c|c}
\midrule
\textbf{\large Model} & \textbf{\large Iteration} & \textbf{\large Time (h)} & \textbf{\large 2.5-bit} & \textbf{\large 3-bit} & \textbf{\large 3.5-bit} & \textbf{\large 4-bit} \\ \midrule
\multirow{4}{*}{\textbf{\large 7B}} & 100 & 2 & 12.34 & 9.10 & 7.90 & 7.55 \\
 & 200 & 5 & 12.37 & 9.03 & 7.90 & 7.54 \\
 & 300 & 11 & 12.32 & 9.06 & 7.89 & 7.53 \\
 & 400 & 21 & 12.32 & 9.06 & 7.90 & 7.53 \\ \midrule
\multirow{4}{*}{\textbf{\large 13B}} & 100 & 3 & 9.43 & 7.83 & 7.16 & 6.92 \\
 & 200 & 8 & 9.39 & 7.77 & 7.13 & 6.91 \\
 & 300 & 16 & 9.39 & 7.79 & 7.14 & 6.90 \\
 & 400 & 29 & 9.39 & 7.76 & 7.13 & 6.90 \\ \bottomrule \bottomrule
\end{tabular}%
}
\vspace{-0.3cm}
\caption{Search cost and C4 validation set perplexity over different numbers of search iterations of Llama 2 7B/13B. The default iteration for Llama 2 7B/13B is 200.}
\vspace{-0.6cm}
\label{tab:iter_result}
\end{table}

\textbf{Search Method.}
NSGA-II is widely used in space exploration studies~\cite{deepmaker, nsga_net, multi_obj_nas} as a standard multi-objective genetic algorithm. While other approaches such as NSGA-III~\cite{nsga3} and MOEAD~\cite{moead} exist, they often extend NSGA-II or require additional hyperparameters, such as reference directions. We therefore adopt NSGA-II for its simplicity and effectiveness.

Although single-objective optimization methods, such as genetic algorithms, reinforcement learning, and policy gradients, may be suitable when optimizing for accuracy at a fixed average bit-width, our goal is to identify the Pareto frontier. This makes NSGA-II the most appropriate choice for our setting.

\textbf{Quantization Proxy.}
For the quantization proxy, we considered data-independent methods like RTN. However, they either yield lower performance or offer no clear advantage in correlation compared to HQQ. Thus, we adopt HQQ for its stronger alignment with actual performance metrics.

\textbf{Predictor.}
We also explored alternative predictor models, including multilayer perceptrons (MLPs), classification and regression trees (CART), and Gaussian processes. Due to slower training or negligible performance improvements over radial basis function (RBF) models, we selected RBF predictors for their efficiency. \Cref{tab:mlp_rbf_predictor} shows a comparison of AMQ using MLP and RBF predictors on the Llama 2 7B model, revealing minimal performance differences between the two.

\textbf{Iteration.}
Table \ref{tab:iter_result} reports C4 perplexity and search time for Llama 2 7B/13B under varying iterations. Our search-and-update procedure considers only Pareto-superior candidates, ensuring stable convergence. However, increasing the iteration limit substantially raises costs, primarily due to predictor training and search, while offering negligible performance improvements. Accordingly, we set the iteration limit to 200 for models up to 30B and 250 for larger models. Nevertheless, when resources or time are limited, using only 100 iterations still produces competitive models, demonstrating that AMQ is a practical rather than time-consuming method.

\section{Additional Visualization of Bit Allocation}
\begin{figure}[t]
\centering
\includegraphics[width=\columnwidth]{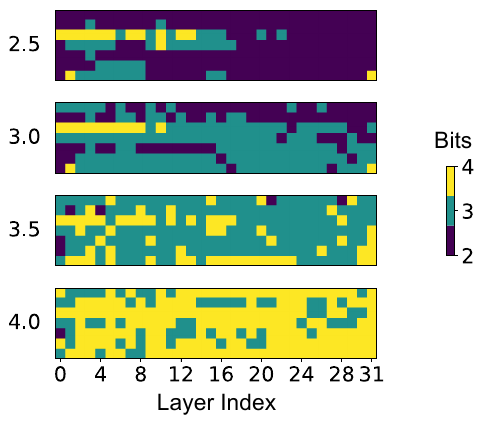}
\vspace{-0.5cm}
\caption{Visualization of bit allocation over linear layers with different average bits at Llama 2 7B. Each row in each box represents Q, K, V, O, Gate, Up, and Down. The numbers on the left indicate the average bits per configuration.}
\vspace{-0.4cm}
\label{fig:bit_assign_7b}
\end{figure}

\begin{figure*}[t]
\centering
\includegraphics[width=\textwidth]{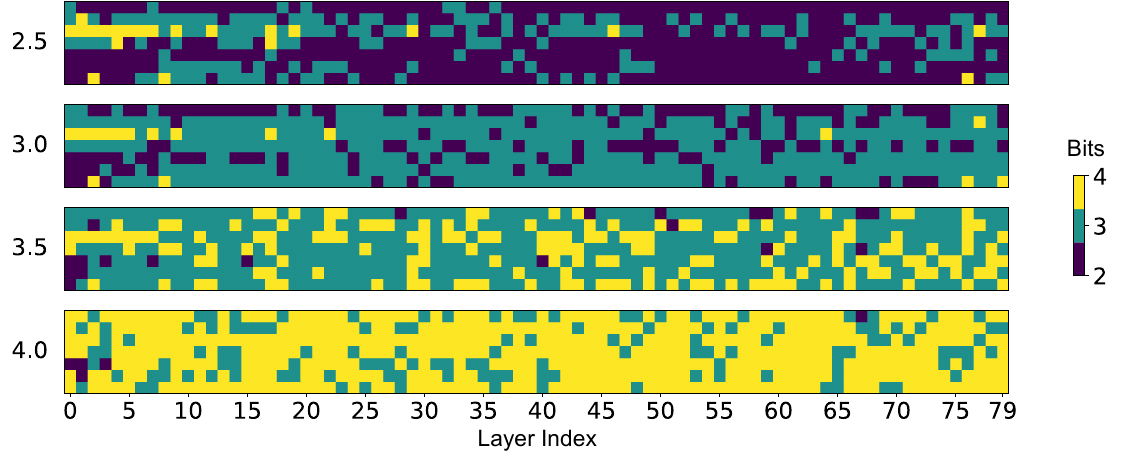}
\vspace{-0.85cm}
\caption{Visualization of bit allocation over linear layers with different average bits at Llama 2 70B. Each row in each box represents Q, K, V, O, Gate, Up, and Down. The numbers on the left indicate the average bits per configuration.}
\vspace{-0.4cm}
\label{fig:bit_assign_70b}
\end{figure*}

\Cref{fig:bit_assign_7b} and \Cref{fig:bit_assign_70b} visualize the bit configuration on various bit-widths searched from AMQ with Llama 2 7B/70B.

\section{Comparison with Other Discrete Structure Search Methods}
Given the vast search space of all possible bit-width assignments for Llama 2 7B ($3^{224}\approx10^{106}$), exhaustive grid search is infeasible within a practical time. For this reason, we propose two lightweight search methods.

\begin{itemize}
\item \textbf{One-shot search.} Layers are first ranked by JSD sensitivity, then the most sensitive layers are assigned 4 bits and the least sensitive layers 2 bits so as to match a target average bit-width in one pass.
\item \textbf{Greedy search.} Starting from all layers at 4 bits, we iteratively quantize one layer to 2 bits, measure the impact on JSD loss, and permanently fix the layer causing the smallest quality drop. This repeats until the target average bit-width is reached.
\end{itemize}

\Cref{tab:one_shot_greedy_cost} and \Cref{tab:one_shot_greedy_search} demonstrate that AMQ outperformed both one-shot and greedy search methods, resulting in significantly lower perplexity and higher zero-shot task performance. These results highlight the effectiveness of AMQ’s finer-grained approach, which surpasses heuristic methods in optimizing the trade-off between memory and quality.

\renewcommand{\arraystretch}{0.5}
\begin{table}[tb!]
\centering
\begin{tabular}{c|c|c}
\midrule
\multirow{3}{*}{\textbf{Method}} & \multicolumn{2}{c}{\textbf{Cost (h)}} \\ \cmidrule{2-3}
 & \textbf{7B} & \textbf{13B} \\ \midrule
One-shot search & 0.1 & 0.3 \\ \midrule
Greedy search & 10 & 43 \\ \midrule
\rowcolor{gray!25} AMQ & 7 & 16 \\ \bottomrule \bottomrule
\end{tabular}
\caption{Cost of one-shot search and greedy search over Llama 2 7B/13B on single NVIDIA RTX6000-ADA.}
\label{tab:one_shot_greedy_cost}
\end{table}

\renewcommand{\arraystretch}{0.75}
\begin{table*}[tb!]
\resizebox{\textwidth}{!}{
\begin{tabular}{c|c|c|c||cc||ccccccc}
\midrule
\textbf{\large Model} & \textbf{\begin{tabular}[c]{@{}c@{}}\textbf{\large Memory}\\ \textbf{\large (MB)}\end{tabular}} & \begin{tabular}[c]{@{}c@{}}\textbf{\large Average}\\ \textbf{\large Bits}\end{tabular} & \textbf{\large Method} & \textbf{\large Wiki2($\downarrow$)} & \textbf{\large C4($\downarrow$)} & \textbf{\large ARC-e($\uparrow$)} & \textbf{\large ARC-c($\uparrow$)} & \textbf{\large PIQA($\uparrow$)} & \textbf{\large HellaS.($\uparrow$)} & \textbf{\large WinoG.($\uparrow$)} & \textbf{\large BoolQ($\uparrow$)} & \textbf{\large Avg.($\uparrow$)} \\ \midrule
\multirow{11}{*}{\textbf{\large 7B}} & 12,853 & 16 & FP16 & 5.47 & 7.26 & 74.58 & 46.25 & 79.11 & 76.00 & 69.22 & 77.77 & 70.49 \\ \cmidrule{2-13}
 & \multirow{3}{*}{2,431} & \multirow{3}{*}{2.5} & One-shot & 10.22 & 13.46 & 58.75 & 34.30 & 72.20 & 62.31 & 61.09 & 65.11 & 58.96 \\
 &  &  & Greedy & 9.98 & 13.16 & 57.15 & 35.07 & 72.91 & 63.94 & 62.67 & \textbf{66.94} & 59.78 \\
 \rowcolor{gray!25} \cellcolor{white} & \cellcolor{white} & \cellcolor{white} & AMQ & \textbf{9.24} & \textbf{12.37} & \textbf{58.88} & \textbf{36.86} & \textbf{73.50} & \textbf{65.01} & \textbf{62.75} & 66.39 & \textbf{60.56} \\ \cmidrule{2-13}
 & \multirow{3}{*}{2,817} & \multirow{3}{*}{3.0} & One-shot & 8.04 & 10.75 & 59.93 & 37.71 & 74.70 & 66.87 & 63.69 & 65.20 & 61.35 \\
 &  &  & Greedy & 7.72 & 10.33 & 64.27 & 38.14 & 75.24 & 68.05 & 64.96 & 64.77 & 62.57 \\
 \rowcolor{gray!25} \cellcolor{white} & \cellcolor{white} & \cellcolor{white} & AMQ & \textbf{6.83} & \textbf{9.03} & \textbf{68.22} & \textbf{41.72} & \textbf{76.55} & \textbf{71.27} & \textbf{67.32} & \textbf{68.44} & \textbf{65.59} \\ \cmidrule{2-13}
 & \multirow{3}{*}{3,203} & \multirow{3}{*}{3.5} & One-shot & 6.77 & 8.96 & 67.68 & 41.47 & 76.66 & 70.97 & 66.61 & 67.55 & 65.16 \\
 &  &  & Greedy & 6.75 & 8.92 & 68.31 & 41.89 & 76.99 & 70.92 & 67.32 & 67.16 & 65.43 \\
 \rowcolor{gray!25} \cellcolor{white} & \cellcolor{white} & \cellcolor{white} & AMQ & \textbf{5.95} & \textbf{7.90} & \textbf{71.55} & \textbf{44.20} & \textbf{77.86} & \textbf{73.92} & \textbf{69.06} & \textbf{73.88} & \textbf{68.41} \\ \midrule
\multirow{12}{*}{\textbf{\large 13B}} & 24,826 & 16 & FP16 & 4.88 & 6.73 & 77.53 & 49.15 & 80.52 & 79.37 & 72.30 & 80.55 & 73.23 \\ \cmidrule{2-13}
 & \multirow{3}{*}{4,408} & \multirow{3}{*}{2.5} & One-shot & 7.74 & 10.54 & 65.24 & 39.51 & 75.03 & 68.37 & 66.46 & 74.16 & 64.79 \\
 &  &  & Greedy & 7.17 & 9.71 & \textbf{67.80} & \textbf{40.44} & 76.01 & 69.68 & \textbf{69.69} & 76.02 & 66.61 \\
 \rowcolor{gray!25} \cellcolor{white} & \cellcolor{white} & \cellcolor{white} & AMQ & \textbf{6.88} & \textbf{9.46} & \textbf{67.38} & 40.19 & \textbf{77.09} & \textbf{71.11} & 69.38 & \textbf{77.16} & \textbf{67.05} \\ \cmidrule{2-13}
 & \multirow{3}{*}{5,164} & \multirow{3}{*}{3.0} & One-shot & 6.56 & 8.89 & 70.29 & 41.98 & 77.69 & 71.62 & 69.30 & 75.44 & 67.72 \\
 &  &  & Greedy & 6.32 & 8.66 & 70.88 & 43.69 & 76.99 & 72.35 & 69.46 & 78.62 & 68.66 \\
 \rowcolor{gray!25} \cellcolor{white} & \cellcolor{white} & \cellcolor{white} & AMQ & \textbf{5.68} & \textbf{7.80} & 72.18 & \textbf{45.39} & \textbf{78.35} & \textbf{76.02} & \textbf{70.32} & \textbf{79.57} & \textbf{70.31} \\ \cmidrule{2-13}
 & \multirow{3}{*}{5,920} & \multirow{3}{*}{3.5} & One-shot & 5.80 & 7.80 & 71.51 & 44.62 & 78.94 & 74.26 & 71.51 & 77.68 & 69.75 \\
 &  &  & Greedy & 5.74 & 7.81 & 72.39 & 45.05 & 79.54 & 75.00 & 70.96 & 78.81 & 70.29 \\
 \rowcolor{gray!25} \cellcolor{white} & \cellcolor{white} & \cellcolor{white} & AMQ & \textbf{5.20} & \textbf{7.13} & \textbf{75.84} & \textbf{48.89} & \textbf{80.25} & \textbf{77.67} & \textbf{72.22} & \textbf{79.14} & \textbf{72.34} \\ \bottomrule \bottomrule
\end{tabular}
}
\caption{Evaluation of one-shot search, greedy search and AMQ on Llama 2 7B/13B.}
\label{tab:one_shot_greedy_search}
\end{table*}

\section{Additional Experiments}
\label{additional_expr}
\Cref{tab:Llama_31_any-size compression}, \Cref{tab:Llama-31-fixed-precision-quantization}, \Cref{tab:Qwen_25_any-size compression}, \Cref{tab:Qwen_25_fixed-precision-quantization}, \Cref{tab:Mistral_7B_v0.3_any-size compression}, and \Cref{tab:Mistral_7B_v0.3_Bit-Targeted-Compression} are the evaluations of Llama 3.1 8/70B, Qwen2.5 7/14B, Mistral 7B v0.3 with AMQ and baselines, respectively.

\begin{table*}[ht]
\resizebox{\textwidth}{!}{
\begin{tabular}{c|c|c|c||cc||ccccccc}
\midrule
\textbf{\large Model} & \textbf{\begin{tabular}[c]{@{}c@{}}\textbf{\large Memory}\\ \textbf{\large (MB)}\end{tabular}} & \begin{tabular}[c]{@{}c@{}}\textbf{\large Average}\\ \textbf{\large Bits}\end{tabular} & \textbf{\large Method} & \textbf{\large Wiki2($\downarrow$)} & \textbf{\large C4($\downarrow$)} & \textbf{\large ARC-e($\uparrow$)} & \textbf{\large ARC-c($\uparrow$)} & \textbf{\large PIQA($\uparrow$)} & \textbf{\large HellaS.($\uparrow$)} & \textbf{\large WinoG.($\uparrow$)} & \textbf{\large BoolQ($\uparrow$)} & \textbf{\large Avg.($\uparrow$)} \\ \midrule
& 12,853 & 16 & FP16 & 5.47 & 7.26 & 74.58 & 46.25 & 79.11 & 76.00 & 69.22 & 77.77 & 70.49 \\ \cmidrule{2-13}
 & \multirow{2}{*}{2,238} & \multirow{2}{*}{2.25} & GPTQ$_{w2g128}$ & 61.77 & 44.10 & 35.40 & 25.17 & 58.32 & 40.17 & 51.46 & 48.62 & 43.19 \\
 &  &  & AWQ$_{w2g128}$ & 2.22e5 & 1.68e5 & 25.76 & 26.62 & 50.38 & 26.07 & 50.04 & 37.83 & 36.12 \\
 \rowcolor{gray!25} \cellcolor{white}& \cellcolor{white}2,315 & \cellcolor{white}2.35 & AMQ & \textbf{11.49} & \textbf{15.12} & \textbf{54.92} & \textbf{33.96} & \textbf{71.06} & \textbf{60.98} & \textbf{60.93} & \textbf{65.32} & \textbf{57.86} \\ \cmidrule{2-13}
 & \multirow{3}{*}{2,817} & \multirow{3}{*}{3.0} & GPTQ$_{w3}$ & 9.27 & 11.81 & 57.62 & 34.73 & 74.43 & 65.68 & 62.67 & \textbf{69.05} & 60.70 \\
 &  &  & AWQ$_{w3}$ & 15.45 & 17.44 & 53.54 & 33.53 & 66.21 & 56.53 & 60.69 & 57.52 & 54.67 \\
 \rowcolor{gray!25} \cellcolor{white} & \cellcolor{white} & \cellcolor{white} &  AMQ & \textbf{6.83} & \textbf{9.03} & \textbf{68.22} & \textbf{41.72} & \textbf{76.55} & \textbf{71.27} & \textbf{67.32} & 68.44 & \textbf{65.59} \\ \cmidrule{2-13}
 & \multirow{3}{*}{3,010} & \multirow{3}{*}{3.25} & GPTQ$_{w3g128}$ & 6.45 & 8.53 & \textbf{69.70} & 43.09 & 77.53 & 71.94 & 68.11 & 72.97 & 67.22 \\
 &  &  & AWQ$_{w3g128}$ & 6.25 & 8.30 & 69.53 & \textbf{44.20} & 77.48 & \textbf{73.33} & 68.11 & 73.12 & 67.63 \\
 \rowcolor{gray!25} \cellcolor{white} & \cellcolor{white} & \cellcolor{white} &  AMQ & \textbf{6.20} & \textbf{8.25} & 69.49 & 43.34 & \textbf{78.13} & 73.21 & \textbf{68.90} & \textbf{74.56} & \textbf{67.94} \\ \cmidrule{2-13}
 & \multirow{3}{*}{3,589} & \multirow{3}{*}{4.0} & GPTQ$_{w4}$ & 6.09 & 7.86 & 71.76 & 43.69 & 77.75 & 74.56 & \textbf{68.90} & 74.65 & 68.55 \\
 &  &  & AWQ$_{w4}$ & 5.83 & 7.72 & 70.75 & 44.11 & 78.13 & 74.93 & 68.67 & \textbf{78.01} & 69.10 \\
 \rowcolor{gray!25} \cellcolor{white}\multirow{-16}{*}{\textbf{\large 7B}} & \cellcolor{white} & \cellcolor{white} &  AMQ & \textbf{5.68} & \textbf{7.54} & \textbf{72.10} & \textbf{44.54} & \textbf{78.45} & \textbf{74.99} & 68.03 & 77.61 & \textbf{69.29} \\ \midrule
 & 24,826 & 16 & FP16 & 4.88 & 6.73 & 77.53 & 49.15 & 80.52 & 79.37 & 72.30 & 80.55 & 73.23 \\  \cmidrule{2-13}
 & \multirow{2}{*}{4,029} & \multirow{2}{*}{2.25} & GPTQ$_{w2g128}$ & 27.78 & 23.39 & 40.57 & 25.68 & 62.08 & 42.37 & 52.17 & 50.98 & 45.64 \\
 &  &  & AWQ$_{w2g128}$ & 1.22e5 & 9.55e4 & 27.10 & 27.47 & 49.95 & 25.99 & 50.83 & 62.17 & 40.59 \\
 \rowcolor{gray!25} \cellcolor{white} & \cellcolor{white}4,181 & \cellcolor{white}2.35 &  AMQ & \textbf{7.60} & \textbf{10.29} & \textbf{65.28} & \textbf{38.99} & \textbf{75.19} & \textbf{67.92} & \textbf{65.98} & \textbf{74.16} & \textbf{64.59} \\ \cmidrule{2-13}
 & \multirow{3}{*}{5,164} & \multirow{3}{*}{3.0} & GPTQ$_{w3}$ & 6.75 & 8.96 & 68.10 & 41.38 & 76.99 & 71.40 & 69.69 & 76.76 & 67.39 \\
 &  &  & AWQ$_{w3}$ & 6.45 & 9.07 & 70.58 & 45.22 & 77.97 & 72.62 & 65.11 & 72.54 & 67.34 \\
 \rowcolor{gray!25} \cellcolor{white} & \cellcolor{white} & \cellcolor{white} &  AMQ & \textbf{5.68} & \textbf{7.80} & \textbf{72.18} & \textbf{45.39} & \textbf{78.35} & \textbf{76.02} & \textbf{70.32} & \textbf{79.57} & \textbf{70.31} \\ \cmidrule{2-13}
 & \multirow{3}{*}{5,542} & \multirow{3}{*}{3.25} & GPTQ$_{w3g128}$ & 5.48 & 7.49 & 74.54 & 46.93 & \textbf{79.22} & 76.83 & 69.38 & 78.44 & 70.89 \\
 &  &  & AWQ$_{w3g128}$ & \textbf{5.32} & \textbf{7.31} & \textbf{75.38} & \textbf{49.06} & 78.94 & \textbf{77.41} & \textbf{72.06} & \textbf{79.82} & \textbf{72.11} \\
 \rowcolor{gray!25} \cellcolor{white} & \cellcolor{white} & \cellcolor{white} &  AMQ & 5.36 & 7.33 & 74.96 & 47.27 & 79.00 & 77.18 & 71.43 & 79.27 & 71.52 \\ \cmidrule{2-13}
 & \multirow{3}{*}{6,676} & \multirow{3}{*}{4.0} & GPTQ$_{w4}$ & 5.19 & 7.06 & 75.04 & 47.10 & \textbf{80.25} & 78.39 & 71.35 & 78.99 & 71.85 \\
 &  &  & AWQ$_{w4}$ & 5.06 & 6.96 & \textbf{77.40} & \textbf{49.40} & 79.65 & 78.57 & 71.90 & 78.59 & 72.59 \\
 \rowcolor{gray!25} \cellcolor{white}\multirow{-16}{*}{\textbf{\large 13B}} & \cellcolor{white} & \cellcolor{white} &  AMQ & \textbf{5.03} & \textbf{6.91} & 77.02 & 48.63 & \textbf{80.25} & \textbf{78.84} & \textbf{72.93} & \textbf{80.18} & \textbf{72.98} \\ \midrule
& 131,563 & 16 & FP16 & 3.32 & 5.71 & 81.02 & 57.25 & 82.70 & 83.78 & 77.98 & 83.79 & 77.75 \\ \cmidrule{2-13}
 & \multirow{2}{*}{19,363} & \multirow{2}{*}{2.25} & GPTQ$_{w2g128}$ & 8.33 & 10.71 & 55.35 & 35.41 & 72.42 & 64.59 & 67.56 & 67.06 & 60.40 \\
 &  &  & AWQ$_{w2g128}$ & 7.25e4 & 6.56e4 & 25.84 & 28.41 & 49.78 & 25.72 & 51.30 & 62.17 & 40.54 \\
 \rowcolor{gray!25} \cellcolor{white} & \cellcolor{white}20,179 & \cellcolor{white}2.35 &  AMQ & \textbf{5.17} & \textbf{7.59} & \textbf{76.05} & \textbf{50.17} & \textbf{79.82} & \textbf{77.54} & \textbf{74.51} & \textbf{82.63} & \textbf{73.45} \\ \cmidrule{2-13}
 & \multirow{3}{*}{25,483} & \multirow{3}{*}{3.0} & GPTQ$_{w3}$ & 4.88 & 7.11 & 75.76 & 49.66 & 80.85 & 79.47 & \textbf{75.45} & 78.65 & 73.31 \\
 &  &  & AWQ$_{w3}$ & 4.36 & 6.63 & \textbf{80.13} & 55.63 & 80.90 & 80.51 & 73.32 & 80.09 & 75.10 \\
 \rowcolor{gray!25} \cellcolor{white} & \cellcolor{white} & \cellcolor{white} &  AMQ & \textbf{4.01} & \textbf{6.29} & 78.54 & \textbf{55.80} & \textbf{81.77} & \textbf{81.31} & \textbf{75.45} & \textbf{83.46} & \textbf{76.05} \\ \cmidrule{2-13}
 & \multirow{3}{*}{27,523} & \multirow{3}{*}{3.25} & GPTQ$_{w3g128}$ & 3.88 & 6.11 & \textbf{79.67} & 54.69 & \textbf{82.48} & 82.34 & \textbf{77.03} & \textbf{83.61} & \textbf{76.64} \\
 &  &  & AWQ$_{w3g128}$ & 3.74 & 6.04 & 79.63 & \textbf{56.48} & 82.21 & \textbf{82.71} & 75.37 & 83.09 & 76.58 \\
 \rowcolor{gray!25} \cellcolor{white} & \cellcolor{white} & \cellcolor{white} &  AMQ & \textbf{3.73} & \textbf{6.03} & 78.75 & 56.23 & 82.43 & 82.27 & 75.61 & 83.00 & 76.38 \\ \cmidrule{2-13}
 & \multirow{3}{*}{33,643} & \multirow{3}{*}{4.0} & GPTQ$_{w4}$ & 3.59 & 5.90 & 80.22 & 56.31 & 82.64 & 83.10 & 77.19 & 82.97 & 77.07 \\
 &  &  & AWQ$_{w4}$ & 3.48 & 5.84 & \textbf{80.72} & \textbf{56.74} & \textbf{83.03} & 83.27 & 77.27 & 83.43 & 77.41 \\
 \rowcolor{gray!25} \cellcolor{white}\multirow{-16}{*}{\textbf{\large 70B}} & \cellcolor{white} & \cellcolor{white} &  AMQ & \textbf{3.46} & \textbf{5.80} & 79.92 & 56.57 & 82.86 & \textbf{83.31} & \textbf{77.58} & \textbf{84.62} & \textbf{77.48} \\ \bottomrule \bottomrule
\end{tabular}
}
\caption{Evaluation of fixed-precision quantization methods with AMQ over Llama 2 7B/13B/70B.  We omit the memory overhead of additional quantization parameters in GPTQ and AWQ at w3 and w4, since it is negligible.}
\label{tab:Llama-2-fixed-precision-quantization}
\end{table*}

\begin{table*}[ht]
\resizebox{\textwidth}{!}{
\begin{tabular}{c|c|c|c||cc||ccccccc}
\midrule
\textbf{\large Model} & \textbf{\begin{tabular}[c]{@{}c@{}}\textbf{\large Memory}\\ \textbf{\large (MB)}\end{tabular}} & \begin{tabular}[c]{@{}c@{}}\textbf{\large Average}\\ \textbf{\large Bits}\end{tabular} & \textbf{\large Method} & \textbf{\large Wiki2($\downarrow$)} & \textbf{\large C4($\downarrow$)} & \textbf{\large ARC-e($\uparrow$)} & \textbf{\large ARC-c($\uparrow$)} & \textbf{\large PIQA($\uparrow$)} & \textbf{\large HellaS.($\uparrow$)} & \textbf{\large WinoG.($\uparrow$)} & \textbf{\large BoolQ($\uparrow$)} & \textbf{\large Avg.($\uparrow$)} \\ \midrule
& 15,317 & 16 & FP16 & 6.24 & 9.54 & 81.02 & 53.24 & 81.23 & 78.94 & 73.16 & 82.17 & 74.96 \\ \cmidrule{2-13}
 &  & & BitStack & 23.28 & 38.23 & 59.43 & 32.42 & 71.55 & 52.13 & 62.51 & \textbf{71.10} & 58.19 \\
 \rowcolor{gray!25} \cellcolor{white} & \cellcolor{white}\multirow{-2}{*}{4,085}& \cellcolor{white}\multirow{-2}{*}{2.5} &  AMQ & \textbf{17.76} & \textbf{26.32} & \textbf{59.47} & \textbf{36.26} & \textbf{72.52} & \textbf{60.25} & \textbf{64.56} & 68.75 & \textbf{60.30} \\ \cmidrule{2-13}
 & & & BitStack & 12.55 & 20.47 & 68.64 & 39.33 & 75.41 & 63.35 & 65.67 & 74.01 & 64.40 \\
 \rowcolor{gray!25} \cellcolor{white} & \cellcolor{white}\multirow{-2}{*}{4,501} & \cellcolor{white}\multirow{-2}{*}{3.0} & AMQ & \textbf{9.44} & \textbf{14.68} & \textbf{71.84} & \textbf{44.88} & \textbf{77.69} & \textbf{70.80} & \textbf{71.51} & \textbf{79.63} & \textbf{69.39} \\ \cmidrule{2-13}
 & & & BitStack & 9.47 & 15.29 & 74.12 & 43.69 & 77.37 & 68.61 & 68.59 & 79.17 & 68.59 \\
 \rowcolor{gray!25} \cellcolor{white} & \cellcolor{white}\multirow{-2}{*}{4,917} & \cellcolor{white}\multirow{-2}{*}{3.5} & AMQ & \textbf{7.39} & \textbf{11.56} & \textbf{74.71} & \textbf{47.27} & \textbf{79.27} & \textbf{75.99} & \textbf{72.14} & \textbf{80.09} & \textbf{71.58} \\ \cmidrule{2-13}
 & & & BitStack & 8.39 & 13.47 & 76.64 & 47.78 & 78.94 & 71.61 & 69.53 & \textbf{81.19} & 70.95 \\
 \rowcolor{gray!25} \cellcolor{white}\multirow{-12}{*}{\textbf{\large 8B}} & \cellcolor{white}\multirow{-2}{*}{5,333} & \cellcolor{white}\multirow{-2}{*}{4.0} & AMQ & \textbf{6.83} & \textbf{10.60} & \textbf{79.17} & \textbf{52.13} & \textbf{80.25} & \textbf{77.38} & \textbf{74.11} & 80.86 & \textbf{73.98} \\ \midrule
 & 134,571 & 16 & FP16 & 2.81 & 7.11 & 86.70 & 65.02 & 84.22 & 85.07 & 79.40 & 85.35 & 80.96 \\ \cmidrule{2-13}
 & & & BitStack & \textbf{7.55} & 12.92 & \textbf{80.43} & \textbf{54.18} & 80.09 & \textbf{77.19} & 75.53 & 79.63 & \textbf{74.51} \\
 \rowcolor{gray!25} \cellcolor{white} & \cellcolor{white}\multirow{-2}{*}{24,411} & \cellcolor{white}\multirow{-2}{*}{2.5} & AMQ & 7.62 & \textbf{12.14} & 79.50 & 53.50 & \textbf{80.14} & 75.39 & \textbf{75.85} & \textbf{81.62} & 74.33 \\ \cmidrule{2-13}
 & & & BitStack & 6.38 & 11.21 & 81.44 & 56.66 & 81.66 & 79.40 & 76.95 & 81.68 & 76.30 \\
 \rowcolor{gray!25} \cellcolor{white} & \cellcolor{white}\multirow{-2}{*}{28,491} & \cellcolor{white}\multirow{-2}{*}{3.0} & AMQ & \textbf{5.84} & \textbf{9.74} & \textbf{82.28} & \textbf{59.73} & \textbf{82.86} & \textbf{80.40} & \textbf{77.19} & \textbf{84.37} & \textbf{77.81} \\ \cmidrule{2-13}
 & & & BitStack & 5.44 & 9.52 & 83.54 & 59.47 & 83.24 & 81.72 & 77.82 & 83.64 & 78.24 \\
 \rowcolor{gray!25} \cellcolor{white} & \cellcolor{white}\multirow{-2}{*}{32,571} & \cellcolor{white}\multirow{-2}{*}{3.5} & AMQ & \textbf{4.26} & \textbf{8.20} & \textbf{84.05} & \textbf{60.92} & \textbf{83.73} & \textbf{83.10} & \textbf{78.30} & \textbf{84.59} & \textbf{79.11} \\ \cmidrule{2-13}
 & & & BitStack & 4.98 & 8.92 & 84.64 & 61.69 & 83.19 & 82.01 & \textbf{79.79} & 83.73 & 79.17 \\
 \rowcolor{gray!25} \cellcolor{white}\multirow{-11}{*}{\textbf{\large 70B}} & \cellcolor{white}\multirow{-2}{*}{36,651} & \cellcolor{white}\multirow{-2}{*}{4.0} & AMQ & \textbf{3.49} & \textbf{7.61} & \textbf{85.77} & \textbf{62.80} & \textbf{84.11} & \textbf{84.12} & 78.77 & \textbf{85.26} & \textbf{80.14} \\ \bottomrule \bottomrule
\end{tabular}
}
\caption{Evaluation of any-size compression method with AMQ over Llama 3.1 8B/70B.}
\label{tab:Llama_31_any-size compression}
\end{table*}

\begin{table*}[ht]
\resizebox{\textwidth}{!}{
\begin{tabular}{c|c|c|c||cc||ccccccc}
\midrule
\textbf{\large Model} & \textbf{\begin{tabular}[c]{@{}c@{}}\textbf{\large Memory}\\ \textbf{\large (MB)}\end{tabular}} & \begin{tabular}[c]{@{}c@{}}\textbf{\large Average}\\ \textbf{\large Bits}\end{tabular} & \textbf{\large Method} & \textbf{\large Wiki2($\downarrow$)} & \textbf{\large C4($\downarrow$)} & \textbf{\large ARC-e($\uparrow$)} & \textbf{\large ARC-c($\uparrow$)} & \textbf{\large PIQA($\uparrow$)} & \textbf{\large HellaS.($\uparrow$)} & \textbf{\large WinoG.($\uparrow$)} & \textbf{\large BoolQ($\uparrow$)} & \textbf{\large Avg.($\uparrow$)} \\ \midrule
 & 15,317 & 16 & FP16 & 6.24 & 9.54 & 81.02 & 53.24 & 81.23 & 78.94 & 73.16 & 82.17 & 74.96 \\ \cmidrule{2-13}
 & \multirow{2}{*}{3,877} & \multirow{2}{*}{2.25} & GPTQ$_{w2g128}$ & 3247.77 & 734.82 & 27.31 & 23.55 & 52.07 & 26.85 & 51.38 & 43.61 & 37.46 \\
 &  &  & AWQ$_{w2g128}$ & 1.5.E+06 & 1.9.E+06 & 24.83 & 24.40 & 50.22 & 26.46 & 49.80 & 37.83 & 35.59 \\
 \rowcolor{gray!25} \cellcolor{white} & \cellcolor{white}3,961 & \cellcolor{white}2.35 & AMQ & \textbf{50.00} & \textbf{61.40} & \textbf{47.81} & \textbf{28.41} & \textbf{65.51} & \textbf{45.03} & \textbf{56.75} & \textbf{46.70} & \textbf{48.37} \\ \cmidrule{2-13}
 & \multirow{3}{*}{4,501} & \multirow{3}{*}{3.0} & GPTQ$_{w3}$ & 13.37 & 18.36 & 60.98 & 38.82 & 73.67 & 67.51 & 57.06 & 51.19 & 58.21 \\
 &  &  & AWQ$_{w3}$ & 18.13 & 31.70 & 67.09 & 44.28 & 73.78 & 68.85 & 58.80 & 65.84 & 63.11 \\
 \rowcolor{gray!25} \cellcolor{white} & \cellcolor{white} & \cellcolor{white} & AMQ & \textbf{9.44} & \textbf{14.68} & \textbf{71.84} & \textbf{44.88} & \textbf{77.69} & \textbf{70.80} & \textbf{71.51} & \textbf{79.63} & \textbf{69.39} \\ \cmidrule{2-13}
 & \multirow{3}{*}{4,709} & \multirow{3}{*}{3.25} & GPTQ$_{w3g128}$ & 26.95 & 21.35 & 56.52 & 34.90 & 71.22 & 67.05 & 67.72 & 69.20 & 61.10 \\
 &  &  & AWQ$_{w3g128}$ & 8.14 & 12.79 & 73.91 & \textbf{47.87} & 78.24 & 73.82 & 70.17 & \textbf{79.36} & 70.56 \\
 \rowcolor{gray!25} \cellcolor{white} & \cellcolor{white} & \cellcolor{white} & AMQ & \textbf{7.96} & \textbf{12.45} & \textbf{75.34} & 47.53 & \textbf{78.89} & \textbf{74.39} & \textbf{71.51} & 78.23 & \textbf{70.98} \\ \cmidrule{2-13}
 & \multirow{3}{*}{5,333} & \multirow{3}{*}{4.0} & GPTQ$_{w4}$ & 87.50 & 53.10 & 55.51 & 37.37 & 59.36 & 42.72 & 67.80 & 64.04 & 54.47 \\
 &  &  & AWQ$_{w4}$ & 7.18 & 11.07 & 76.94 & 51.11 & \textbf{80.63} & \textbf{77.52} & 73.32 & 80.73 & 73.38 \\
 \rowcolor{gray!25} \cellcolor{white}\multirow{-16}{*}{\textbf{\large 8B}} & \cellcolor{white} & \cellcolor{white} & AMQ & \textbf{6.83} & \textbf{10.60} & \textbf{79.17} & \textbf{52.13} & 80.25 & 77.38 & \textbf{74.11} & \textbf{80.86} & \textbf{73.98} \\ \midrule
 & 134,571 & 16 & FP16 & 2.81 & 7.11 & 86.70 & 65.02 & 84.22 & 85.07 & 79.40 & 85.35 & 80.96 \\ \cmidrule{2-13}
 & \multirow{2}{*}{22,371} & \multirow{2}{*}{2.25} & GPTQ$_{w2g128}$ & 113.22 & 131.90 & 25.38 & 25.85 & 51.69 & 37.16 & 52.64 & 47.40 & 40.02 \\
 &  &  & AWQ$_{w2g128}$ & 1.8.E+06 & 1.5.E+06 & 24.54 & 26.02 & 51.52 & 26.43 & 53.20 & 62.17 & 40.65 \\
  \rowcolor{gray!25} \cellcolor{white} & \cellcolor{white}23,187 & \cellcolor{white}2.35 & AMQ & \textbf{8.46} & \textbf{13.18} & \textbf{73.91} & \textbf{48.38} & \textbf{78.18} & \textbf{72.79} & \textbf{73.16} & \textbf{79.63} & \textbf{71.01} \\ \cmidrule{2-13}
 & \multirow{3}{*}{28,491} & \multirow{3}{*}{3.0} & GPTQ$_{w3}$ & 1.6.E+04 & 1.3.E+04 & 25.80 & 25.94 & 52.23 & 26.45 & 48.78 & 37.83 & 36.17 \\
 &  &  & AWQ$_{w3}$ & 43.14 & 43.59 & 42.30 & 28.92 & 63.93 & 44.57 & 53.04 & 53.33 & 47.68 \\
 \rowcolor{gray!25} \cellcolor{white} & \cellcolor{white} & \cellcolor{white} & AMQ & \textbf{5.84} & \textbf{9.74} & \textbf{82.28} & \textbf{59.73} & \textbf{82.86} & \textbf{80.40} & \textbf{77.19} & \textbf{84.37} & \textbf{77.81} \\ \cmidrule{2-13}
 & \multirow{3}{*}{30,531} & \multirow{3}{*}{3.25} & GPTQ$_{w3g128}$ & 5.17 & 8.76 & 68.22 & 43.86 & 74.37 & 81.61 & 76.09 & 82.39 & 71.09 \\
 &  &  & AWQ$_{w3g128}$ & \textbf{4.80} & \textbf{8.62} & \textbf{83.96} & \textbf{62.37} & 83.41 & \textbf{82.67} & \textbf{78.85} & 83.64 & \textbf{79.15} \\
 \rowcolor{gray!25} \cellcolor{white} & \cellcolor{white} & \cellcolor{white} & AMQ & 5.09 & 8.91 & 82.95 & 60.67 & \textbf{83.68} & 82.41 & 78.06 & \textbf{85.23} & 78.83 \\ \cmidrule{2-13}
 & \multirow{3}{*}{36,651} & \multirow{3}{*}{4.0} & GPTQ$_{w4}$ & 1.4.E+04 & 8.8.E+03 & 25.29 & 26.79 & 52.12 & 26.43 & 51.85 & 37.86 & 36.73 \\
 &  &  & AWQ$_{w4}$ & 4.18 & 8.29 & 83.00 & 60.32 & 83.19 & 83.39 & 63.06 & 82.75 & 75.95 \\
 \rowcolor{gray!25} \cellcolor{white}\multirow{-16}{*}{\textbf{\large 70B}} & \cellcolor{white} & \cellcolor{white} & AMQ & \textbf{3.49} & \textbf{7.61} & \textbf{85.77} & \textbf{62.80} & \textbf{84.11} & \textbf{84.12} & \textbf{78.77} & \textbf{85.26} & \textbf{80.14} \\ \bottomrule \bottomrule
\end{tabular}
}
\caption{Evaluation of fixed-precision quantization methods with AMQ over Llama 3.1 8B/70B.  We omit the memory overhead of additional quantization parameters in GPTQ and AWQ at w3 and w4, since it is negligible.}
\label{tab:Llama-31-fixed-precision-quantization}
\end{table*}

\begin{table*}[ht]
\resizebox{\textwidth}{!}{
\begin{tabular}{c|c|c|c||cc||ccccccc}
\midrule
\textbf{\large Model} & \textbf{\begin{tabular}[c]{@{}c@{}}\textbf{\large Memory}\\ \textbf{\large (MB)}\end{tabular}} & \begin{tabular}[c]{@{}c@{}}\textbf{\large Average}\\ \textbf{\large Bits}\end{tabular} & \textbf{\large Method} & \textbf{\large Wiki2($\downarrow$)} & \textbf{\large C4($\downarrow$)} & \textbf{\large ARC-e($\uparrow$)} & \textbf{\large ARC-c($\uparrow$)} & \textbf{\large PIQA($\uparrow$)} & \textbf{\large HellaS.($\uparrow$)} & \textbf{\large WinoG.($\uparrow$)} & \textbf{\large BoolQ($\uparrow$)} & \textbf{\large Avg.($\uparrow$)} \\ \midrule
 & 14,525 & 16 & FP16 & 6.85 & 11.89 & 77.69 & 51.45 & 79.92 & 78.96 & 73.01 & 84.56 & 74.26 \\ \cmidrule{2-13}
 &  &  & BitStack & 20.97 & 38.16 & 65.66 & 37.29 & 71.87 & 54.82 & \textbf{62.90} & \textbf{75.41} & 61.33 \\
 \rowcolor{gray!25}\cellcolor{white} & \cellcolor{white}\multirow{-2}{*}{4,025} & \cellcolor{white}\multirow{-2}{*}{2.5} &  AMQ & \textbf{12.85} & \textbf{19.81} & \textbf{67.42} & \textbf{42.24} & \textbf{74.43} & \textbf{66.42} & 61.64 & 73.52 & \textbf{64.28} \\ \cmidrule{2-13}
 &  &  & BitStack & 11.92 & 20.31 & \textbf{75.67} & \textbf{47.78} & 76.01 & 65.48 & 66.46 & 77.58 & 68.16 \\
 \rowcolor{gray!25}\cellcolor{white} & \cellcolor{white}\multirow{-2}{*}{4,414} & \cellcolor{white}\multirow{-2}{*}{3.0} &  AMQ & \textbf{8.74} & \textbf{14.30} & 73.74 & 46.59 & \textbf{78.45} & \textbf{73.88} & \textbf{66.93} & \textbf{82.02} & \textbf{70.27} \\ \cmidrule{2-13}
 &  &  & BitStack & 9.17 & 15.86 & \textbf{79.12} & \textbf{52.13} & 78.94 & 70.22 & \textbf{71.35} & 83.98 & \textbf{72.62} \\
 \rowcolor{gray!25}\cellcolor{white} & \cellcolor{white}\multirow{-2}{*}{4,803} & \cellcolor{white}\multirow{-2}{*}{3.5} &  AMQ & \textbf{7.57} & \textbf{12.78} & 73.36 & 48.04 & \textbf{79.22} & \textbf{76.61} & 70.09 & \textbf{84.07} & 71.90 \\ \cmidrule{2-13}
 &  &  & BitStack & 8.36 & 14.43 & \textbf{79.55} & \textbf{52.39} & 79.60 & 72.43 & \textbf{73.16} & \textbf{84.80} & 73.65 \\
 \rowcolor{gray!25}\cellcolor{white}\multirow{-12}{*}{\textbf{\large 7B}} & \cellcolor{white}\multirow{-2}{*}{5,192} & \cellcolor{white}\multirow{-2}{*}{4.0} & AMQ & \textbf{7.20} & \textbf{12.33} & 77.40 & 51.79 & \textbf{79.98} & \textbf{77.74} & 71.98 & 83.82 & \textbf{73.79} \\ \midrule
 & 28,171 & 16 & FP16 & 5.29 & 10.35 & 79.38 & 58.96 & 82.37 & 82.90 & 75.93 & 85.32 & 77.48 \\ \cmidrule{2-13}
 &  &  & BitStack & 13.14 & 22.54 & 67.68 & 41.38 & 75.19 & 64.40 & \textbf{70.88} & 70.76 & 65.05 \\
 \rowcolor{gray!25}\cellcolor{white} & \cellcolor{white}\multirow{-2}{*}{6,909} & \cellcolor{white}\multirow{-2}{*}{2.5} &  AMQ & \textbf{10.18} & \textbf{16.38} & \textbf{71.42} & \textbf{46.33} & \textbf{75.52} & \textbf{70.94} & 68.03 & \textbf{71.80} & \textbf{67.34} \\ \cmidrule{2-13}
 &  &  & BitStack & 9.12 & 15.71 & 67.97 & 45.22 & 77.69 & 72.50 & \textbf{75.85} & 75.44 & 69.11 \\
 \rowcolor{gray!25}\cellcolor{white} & \cellcolor{white}\multirow{-2}{*}{7,697} & \cellcolor{white}\multirow{-2}{*}{3.0} &  AMQ & \textbf{7.23} & \textbf{12.31} & \textbf{80.56} & \textbf{54.44} & \textbf{80.41} & \textbf{77.83} & 72.85 & \textbf{81.87} & \textbf{74.66} \\ \cmidrule{2-13}
 &  &  & BitStack & 7.29 & 13.22 & \textbf{80.30} & 53.33 & 79.43 & 75.67 & \textbf{77.27} & 80.49 & 74.42 \\
 \rowcolor{gray!25}\cellcolor{white} & \cellcolor{white}\multirow{-2}{*}{8,484} & \cellcolor{white}\multirow{-2}{*}{3.5} &  AMQ & \textbf{6.27} & \textbf{11.15} & 80.01 & \textbf{55.89} & \textbf{80.58} & \textbf{80.78} & 74.74 & \textbf{85.29} & \textbf{76.21} \\ \cmidrule{2-13}
 &  &  & BitStack & 6.72 & 12.30 & \textbf{82.62} & 56.66 & 79.76 & 77.62 & \textbf{77.11} & 82.29 & 76.01 \\
 \rowcolor{gray!25}\cellcolor{white}\multirow{-12}{*}{\textbf{\large 14B}} & \cellcolor{white}\multirow{-2}{*}{9,272} & \cellcolor{white}\multirow{-2}{*}{4.0} &  AMQ & \textbf{5.81} & \textbf{10.73} & 80.85 & \textbf{59.30} & \textbf{81.77} & \textbf{82.06} & 75.69 & \textbf{84.37} & \textbf{77.34} \\ \bottomrule \bottomrule
\end{tabular}
}
\caption{Evaluation of any-size compression method with AMQ over Qwen2.5 7B/14B.}
\label{tab:Qwen_25_any-size compression}
\end{table*}

\begin{table*}[ht]
\resizebox{\textwidth}{!}{
\begin{tabular}{c|c|c|c||cc||ccccccc}
\midrule
\textbf{\large Model} & \textbf{\begin{tabular}[c]{@{}c@{}}\textbf{\large Memory}\\ \textbf{\large (MB)}\end{tabular}} & \begin{tabular}[c]{@{}c@{}}\textbf{\large Average}\\ \textbf{\large Bits}\end{tabular} & \textbf{\large Method} & \textbf{\large Wiki2($\downarrow$)} & \textbf{\large C4($\downarrow$)} & \textbf{\large ARC-e($\uparrow$)} & \textbf{\large ARC-c($\uparrow$)} & \textbf{\large PIQA($\uparrow$)} & \textbf{\large HellaS.($\uparrow$)} & \textbf{\large WinoG.($\uparrow$)} & \textbf{\large BoolQ($\uparrow$)} & \textbf{\large Avg.($\uparrow$)} \\ \midrule
 & 15,317 & 16 & FP16 & 6.85 & 11.89 & 77.69 & 51.45 & 79.92 & 78.96 & 73.01 & 84.56 & 74.26 \\ \cmidrule{2-13}
 & \multirow{2}{*}{3,830} & \multirow{2}{*}{2.25} & GPTQ$_{w2g128}$ & 57.77 & 55.94 & 32.91 & 27.65 & 57.56 & 41.11 & 51.93 & 47.52 & 43.11 \\
 &  &  & AWQ$_{w2g128}$ & 1.1.E+07 & 1.3.E+07 & 26.26 & 26.54 & 51.36 & 25.93 & 49.72 & 37.83 & 36.27 \\
 \rowcolor{gray!25} \cellcolor{white} & \cellcolor{white}3,908 & \cellcolor{white}2.35 & AMQ & \textbf{15.08} & \textbf{23.57} & \textbf{63.59} & \textbf{39.16} & \textbf{72.69} & \textbf{63.86} & \textbf{61.01} & \textbf{70.46} & \textbf{61.80} \\ \cmidrule{2-13}
 & \multirow{3}{*}{4,414} & \multirow{3}{*}{3.0} & GPTQ$_{w3}$ & 13.37 & 18.36 & 60.98 & 38.82 & 73.67 & 67.51 & 57.06 & 51.19 & 58.21 \\
 &  &  & AWQ$_{w3}$ & 18.13 & 31.70 & 67.09 & 44.28 & 73.78 & 68.85 & 58.80 & 65.84 & 63.11 \\
 \rowcolor{gray!25} \cellcolor{white} & \cellcolor{white} & \cellcolor{white} &  AMQ & \textbf{8.74} & \textbf{14.30} & \textbf{73.74} & \textbf{46.59} & \textbf{78.45} & \textbf{73.88} & \textbf{66.93} & \textbf{82.02} & \textbf{70.27} \\ \cmidrule{2-13}
 & \multirow{3}{*}{4,608} & \multirow{3}{*}{3.25} & GPTQ$_{w3g128}$ & 8.19 & 13.28 & 64.69 & 43.77 & 77.75 & 75.74 & 67.96 & 82.14 & 68.67 \\
 &  &  & AWQ$_{w3g128}$ & 8.03 & 13.47 & \textbf{78.66} & \textbf{49.49} & 78.35 & 75.27 & \textbf{68.43} & \textbf{84.98} & \textbf{72.53} \\
 \rowcolor{gray!25} \cellcolor{white} & \cellcolor{white} & \cellcolor{white} &  AMQ & \textbf{7.91} & \textbf{13.21} & 73.91 & 48.04 & \textbf{79.05} & \textbf{75.81} & 66.93 & 84.13 & 71.31 \\ \cmidrule{2-13}
 &  &  & GPTQ$_{w4}$ & 7.65 & 12.74 & 74.45 & 50.17 & 79.87 & 77.04 & 68.27 & 83.06 & 72.14 \\
 &  &  & AWQ$_{w4}$ & 7.63 & 13.13 & 76.77 & 50.26 & 79.22 & \textbf{77.74} & 70.88 & 83.49 & 73.06 \\
 \rowcolor{gray!25} \cellcolor{white}\multirow{-16}{*}{\textbf{\large 7B}} & \cellcolor{white}\multirow{-3}{*}{5,197} & \cellcolor{white}\multirow{-3}{*}{4.0} &  AMQ & \textbf{7.20} & \textbf{12.33} & \textbf{77.40} & \textbf{51.79} & \textbf{79.98} & \textbf{77.74} & \textbf{71.98} & \textbf{83.82} & \textbf{73.79} \\ \midrule
 & 28,172 & 16 & FP16 & 5.29 & 10.35 & 79.38 & 58.96 & 82.37 & 82.90 & 75.93 & 85.32 & 77.48 \\ \cmidrule{2-13}
 & \multirow{2}{*}{6,515} & \multirow{2}{*}{2.25} & GPTQ$_{w2g128}$ & 39.38 & 46.25 & 33.71 & 25.09 & 58.76 & 40.32 & 49.17 & 56.24 & 43.88 \\
 &  &  & AWQ$_{w2g128}$ & 3.7.E+07 & 3.4.E+07 & 25.08 & 27.05 & 52.12 & 26.20 & 49.25 & 62.17 & 40.31 \\
 \rowcolor{gray!25} \cellcolor{white} & \cellcolor{white}6,673 & \cellcolor{white}2.35 &  AMQ & \textbf{12.45} & \textbf{19.90} & \textbf{64.44} & \textbf{40.44} & \textbf{74.32} & \textbf{65.92} & \textbf{65.27} & \textbf{65.81} & \textbf{62.70} \\ \cmidrule{2-13}
 &  &  & GPTQ$_{w3}$ & 9.81 & 14.68 & 68.56 & 43.34 & 77.20 & 72.05 & 63.06 & 57.65 & 63.64 \\
 &  &  & AWQ$_{w3}$ & 8.53 & 14.04 & 70.83 & 47.70 & 78.94 & 77.56 & 66.22 & 71.28 & 68.76 \\
 \rowcolor{gray!25} \cellcolor{white} & \cellcolor{white}\multirow{-3}{*}{7,697} & \cellcolor{white}\multirow{-3}{*}{3.0} &  AMQ & \textbf{7.23} & \textbf{12.31} & \textbf{80.56} & \textbf{54.44} & \textbf{80.41} & \textbf{77.83} & \textbf{72.85} & \textbf{81.87} & \textbf{74.66} \\ \cmidrule{2-13}
 &  &  & GPTQ$_{w3g128}$ & 6.96 & 11.65 & \textbf{82.37} & \textbf{58.02} & \textbf{80.85} & \textbf{79.77} & 72.22 & \textbf{84.65} & \textbf{76.31} \\
 &  &  & AWQ$_{w3g128}$ & 6.65 & 11.60 & 80.77 & 55.80 & 80.79 & 79.48 & \textbf{75.14} & 83.00 & 75.83 \\
 \rowcolor{gray!25} \cellcolor{white} & \cellcolor{white}\multirow{-3}{*}{8,090} &  \cellcolor{white}\multirow{-3}{*}{3.25} & AMQ & \textbf{6.61} & \textbf{11.49} & 79.71 & 54.78 & 80.58 & 79.76 & 72.53 & 84.46 & 75.30 \\ \cmidrule{2-13}
 &  &  & GPTQ$_{w4}$ & 6.31 & 11.10 & 81.48 & 57.34 & \textbf{81.77} & 81.37 & 74.90 & 84.28 & 76.86 \\
 &  &  & AWQ$_{w4}$ & 6.06 & 11.02 & \textbf{81.57} & 58.02 & 81.45 & \textbf{82.25} & 74.43 & \textbf{85.87} & 77.26 \\
 \rowcolor{gray!25} \cellcolor{white}\multirow{-16}{*}{\textbf{\large 14B}} & \cellcolor{white}\multirow{-3}{*}{9,272} & \cellcolor{white}\multirow{-3}{*}{4.0} &  AMQ & \textbf{5.81} & \textbf{10.73} & 80.85 & \textbf{59.30} & \textbf{81.77} & 82.06 & \textbf{75.69} & 84.37 & \textbf{77.34} \\ \bottomrule \bottomrule
\end{tabular}
}
\caption{Evaluation of fixed-precision quantization method with AMQ over Qwen2.5 7B/14B. We omit the memory overhead of additional quantization parameters in GPTQ and AWQ at w3 and w4, since it is negligible.}
\label{tab:Qwen_25_fixed-precision-quantization}
\end{table*}

\begin{table*}[ht]
\resizebox{\textwidth}{!}{
\begin{tabular}{c|c|c||cc||ccccccc}
\midrule
\textbf{\begin{tabular}[c]{@{}c@{}}\textbf{\large Memory}\\ \textbf{\large (MB)}\end{tabular}} & \begin{tabular}[c]{@{}c@{}}\textbf{\large Average}\\ \textbf{\large Bits}\end{tabular} & \textbf{\large Method} & \textbf{\large Wiki2($\downarrow$)} & \textbf{\large C4($\downarrow$)} & \textbf{\large ARC-e($\uparrow$)} & \textbf{\large ARC-c($\uparrow$)} & \textbf{\large PIQA($\uparrow$)} & \textbf{\large HellaS.($\uparrow$)} & \textbf{\large WinoG.($\uparrow$)} & \textbf{\large BoolQ($\uparrow$)} & \textbf{\large Avg.($\uparrow$)} \\ \midrule
13,825 & 16 & FP16 & 5.32 & 8.48 & 78.37 & 52.30 & 82.43 & 80.42 & 73.88 & 82.11 & 74.92 \\ \midrule
 &  & BitStack & 10.68 & 16.32 & 64.44 & 35.67 & 75.24 & 61.71 & 66.54 & 74.22 & 62.97 \\ 
\rowcolor{gray!25} \cellcolor{white}\multirow{-2}{*}{2,593} & \cellcolor{white}\multirow{-2}{*}{2.5} &  AMQ & \textbf{8.34} & \textbf{12.62} & \textbf{66.41} & \textbf{38.82} & \textbf{77.09} & \textbf{70.19} & \textbf{66.69} & \textbf{79.60} & \textbf{66.47} \\ \midrule
 &  & BitStack & 7.94 & 12.20 & 70.71 & 40.87 & 77.48 & 69.57 & 68.75 & 78.44 & 67.63 \\
\rowcolor{gray!25} \cellcolor{white}\multirow{-2}{*}{3,009} & \cellcolor{white}\multirow{-2}{*}{3} &  AMQ & \textbf{6.46} & \textbf{10.06} & \textbf{73.57} & \textbf{45.48} & \textbf{79.49} & \textbf{76.09} & \textbf{69.38} & \textbf{82.54} & \textbf{71.09} \\ \midrule
 & & BitStack & 6.69 & 10.39 & 73.15 & 44.20 & 79.16 & 73.17 & 70.96 & 78.07 & 69.78 \\
\rowcolor{gray!25} \cellcolor{white}\multirow{-2}{*}{3,425} & \cellcolor{white}\multirow{-2}{*}{3.5} &  AMQ & \textbf{5.69} & \textbf{9.01} & \textbf{75.04} & \textbf{49.15} & \textbf{81.07} & \textbf{79.09} & \textbf{71.98} & \textbf{82.75} & \textbf{73.18} \\ \midrule
 & & BitStack & 6.24 & 9.70 & 75.29 & 46.25 & 79.92 & 75.43 & 70.48 & 80.83 & 71.37 \\
\rowcolor{gray!25} \cellcolor{white}\multirow{-2}{*}{3,841} & \cellcolor{white}\multirow{-2}{*}{4} &  AMQ & \textbf{5.49} & \textbf{8.74} & \textbf{77.36} & \textbf{51.28} & \textbf{81.77} & \textbf{79.71} & \textbf{72.30} & \textbf{82.91} & \textbf{74.22} \\ \bottomrule \bottomrule
\end{tabular}
}
\caption{Evaluation of any-size compression method with AMQ over Mistral 7B v0.3.}
\label{tab:Mistral_7B_v0.3_any-size compression}
\end{table*}

\begin{table*}[ht]
\resizebox{\textwidth}{!}{
\begin{tabular}{c|c|c||cc||ccccccc}
\midrule
\textbf{\begin{tabular}[c]{@{}c@{}}\textbf{\large Memory}\\ \textbf{\large (MB)}\end{tabular}} & \begin{tabular}[c]{@{}c@{}}\textbf{\large Average}\\ \textbf{\large Bits}\end{tabular} & \textbf{\large Method} & \textbf{\large Wiki2($\downarrow$)} & \textbf{\large C4($\downarrow$)} & \textbf{\large ARC-e($\uparrow$)} & \textbf{\large ARC-c($\uparrow$)} & \textbf{\large PIQA($\uparrow$)} & \textbf{\large HellaS.($\uparrow$)} & \textbf{\large WinoG.($\uparrow$)} & \textbf{\large BoolQ($\uparrow$)} & \textbf{\large Avg.($\uparrow$)} \\ \midrule
13,825 & 16 & FP16 & 5.32 & 8.48 & 78.37 & 52.30 & 82.43 & 80.42 & 73.88 & 82.11 & 74.92 \\ \midrule
 \multirow{2}{*}{2,385} & \multirow{2}{*}{2.25} & GPTQ$_{w2g128}$ & 23.71 & 27.85 & 40.32 & 28.84 & 60.07 & 42.68 & 54.62 & 51.04 & 46.26 \\ 
 &  & AWQ$_{w2g128}$ & 3.7.E+04 & 3.7.E+04 & 26.05 & 28.67 & 51.14 & 25.83 & 49.88 & 37.83 & 36.57 \\ 
\rowcolor{gray!25} \cellcolor{white}2,469 & \cellcolor{white}2.35 & AMQ & \textbf{10.34} & \textbf{15.47} & \textbf{64.18} & \textbf{35.58} & \textbf{76.88} & \textbf{65.78} & \textbf{63.93} & \textbf{75.96} & \textbf{63.72} \\ \midrule
\multirow{3}{*}{3,009} & \multirow{3}{*}{3} & GPTQ$_{w3}$ & 9.55 & 13.57 & 66.08 & 42.06 & 77.42 & 72.27 & 63.30 & 68.72 & 64.97 \\
 &  & AWQ$_{w3}$ & 7.54 & 12.14 & 72.56 & 43.94 & 78.94 & 74.49 & 64.48 & 70.46 & 67.48 \\
\rowcolor{gray!25} \cellcolor{white} & \cellcolor{white} &  AMQ & \textbf{6.46} & \textbf{10.06} & \textbf{73.57} & \textbf{45.48} & \textbf{79.49} & \textbf{76.09} & \textbf{69.38} & \textbf{82.54} & \textbf{71.09} \\ \midrule
\multirow{3}{*}{3,217} & \multirow{3}{*}{3.25} & GPTQ$_{w3g128}$ & 6.20 & 9.63 & 73.11 & 46.76 & 79.82 & 77.68 & 71.35 & 78.38 & 71.18 \\
 &  & AWQ$_{w3g128}$ & 5.92 & 9.34 & \textbf{75.29} & \textbf{48.89} & \textbf{80.36} & 77.43 & 71.03 & 79.27 & 72.05 \\
 \rowcolor{gray!25} \cellcolor{white} & \cellcolor{white} &  AMQ & \textbf{5.91} & \textbf{9.33} & 75.08 & \textbf{48.89} & 79.82 & \textbf{78.28} & \textbf{71.74} & \textbf{82.17} & \textbf{72.66} \\ \midrule
\multirow{3}{*}{3,841} & \multirow{3}{*}{4} & GPTQ$_{w4}$ & 5.74 & 9.01 & 75.97 & 49.49 & 80.58 & 77.22 & 71.11 & 80.64 & 72.50 \\
 &  & AWQ$_{w4}$ & 5.72 & 9.02 & 77.02 & 50.60 & 80.63 & 79.20 & \textbf{72.69} & 79.17 & 73.22 \\
 \rowcolor{gray!25} \cellcolor{white} & \cellcolor{white} & AMQ & \textbf{5.49} & \textbf{8.74} & \textbf{77.36} & \textbf{51.28} & \textbf{81.77} & \textbf{79.71} & 72.30 & \textbf{82.91} & \textbf{74.22}  \\ \bottomrule \bottomrule
\end{tabular}
}
\caption{Evaluation of fixed-precision quantization methods with AMQ over Mistral 7B v0.3. We omit the memory overhead of additional quantization parameters in GPTQ and AWQ at w3 and w4, since it is negligible.}
\label{tab:Mistral_7B_v0.3_Bit-Targeted-Compression}
\end{table*}


\end{document}